%% file: anonymous-submission-latex-2025.tex
\documentclass[letterpaper]{article} % DO NOT CHANGE THIS
\usepackage[draft]{aaai25}  % DO NOT CHANGE THIS
\usepackage{times}  % DO NOT CHANGE THIS
\usepackage{helvet}  % DO NOT CHANGE THIS
\usepackage{courier}  % DO NOT CHANGE THIS
\usepackage[hyphens]{url}  % DO NOT CHANGE THIS
\usepackage{graphicx} % DO NOT CHANGE THIS
\usepackage{natbib}  % DO NOT CHANGE THIS AND DO NOT ADD ANY OPTIONS TO IT
\usepackage{caption} % DO NOT CHANGE THIS AND DO NOT ADD ANY OPTIONS TO IT
\usepackage{algorithm}
\usepackage{algorithmic}
\usepackage{multirow}
\usepackage{tabularx}
\usepackage{color}
\usepackage{soul}
\usepackage{subfigure}
\usepackage{amsmath}
\definecolor{blue}{RGB}{0, 0, 255}

\usepackage{newfloat}
\usepackage{listings}
\DeclareCaptionStyle{ruled}{labelfont=normalfont,labelsep=colon,strut=off} % DO NOT CHANGE THIS
\floatstyle{ruled}
\newfloat{listing}{tb}{lst}{}
\floatname{listing}{Listing}
\title{Strategic Chain-of-Thought: Guiding Accurate Reasoning in LLMs through Strategy Elicitation}
\author {
    Yu Wang\textsuperscript{\rm 1,\rm 2},
    Shiwan Zhao\textsuperscript{\rm 3,\rm 4},
    Zhihu Wang\textsuperscript{\rm 1},
    Heyuan Huang\textsuperscript{\rm 1},
    Ming Fan\textsuperscript{\rm 2},
    Yubo Zhang\textsuperscript{\rm 1},\\
    Zhixing Wang\textsuperscript{\rm 1},
    Haijun Wang\textsuperscript{\rm 2},
    Ting Liu\textsuperscript{\rm 2}
}

\affiliations{
    \textsuperscript{\rm 1}Huawei Technologies Ltd.\\
    \textsuperscript{\rm 2}Xi’an Jiaotong University\\
    \textsuperscript{\rm 3}Nankai University\\
    \textsuperscript{\rm 4}Shanghai Jiao Tong University\\
}

\usepackage{bibentry}
\begin{document}

\maketitle

\begin{abstract}
The Chain-of-Thought (CoT) paradigm has emerged as a critical approach for enhancing the reasoning capabilities of large language models (LLMs). 
However, despite their widespread adoption and success, CoT methods often exhibit instability due to their inability to consistently ensure the quality of generated reasoning paths, leading to sub-optimal reasoning performance.
To address this challenge, we propose the \textbf{Strategic Chain-of-Thought} (SCoT), a novel methodology designed to refine LLM performance by integrating strategic knowledge prior to generating intermediate reasoning steps. 
SCoT employs a two-stage approach within a single prompt: first eliciting an effective problem-solving strategy, which is then used to guide the generation of high-quality CoT paths and final answers. Our experiments across eight challenging reasoning datasets demonstrate significant improvements, including a 21.05\% increase on the GSM8K dataset and 24.13\% on the Tracking\_Objects dataset, respectively, using the Llama3-8b model.
Additionally, we extend the SCoT framework to develop a few-shot method with automatically matched demonstrations, yielding even stronger results. These findings underscore the efficacy of SCoT, highlighting its potential to substantially enhance LLM performance in complex reasoning tasks.
\end{abstract}

\section{Introduction}
The rapid development of large language models (LLMs) has highlighted their remarkable effectiveness in reasoning tasks~\cite{huang2022towards, chang2024survey}, particularly when integrated with various prompting techniques~\cite{sivarajkumar2023empiricalevaluationpromptingstrategies}. These techniques consistently enable impressive performance across diverse domains. Among them, the Chain-of-Thought (CoT) paradigm has played a pivotal role in enhancing the reasoning capabilities of LLMs~\cite{ kojima2022large,zhang2022automatic,wang2023plan}.
As a result, CoT has become a fundamental component of contemporary LLMs and is now widely adopted in the field of natural language processing. 

Despite the demonstrated effectiveness of the CoT approach in various applications, it faces significant challenges in complex reasoning tasks. These challenges primarily arise from the variability in the quality of the reasoning paths generated by the CoT method~\cite{wang2022self}, which are not consistently optimal. Consequently, even when LLMs produce a CoT path that aligns with a valid reasoning process, there remains a risk that the final outcome may be erroneous.

This phenomenon is analogous to findings in cognitive science, where different problem-solving strategies, although correct, can vary in their likelihood of producing errors. According to Sweller's Cognitive Load Theory~\cite{swell}, different problem-solving strategies impose varying levels of cognitive load, leading to different probabilities of error.
\begin{figure}[t]
\centering
\includegraphics[width=0.48\textwidth]{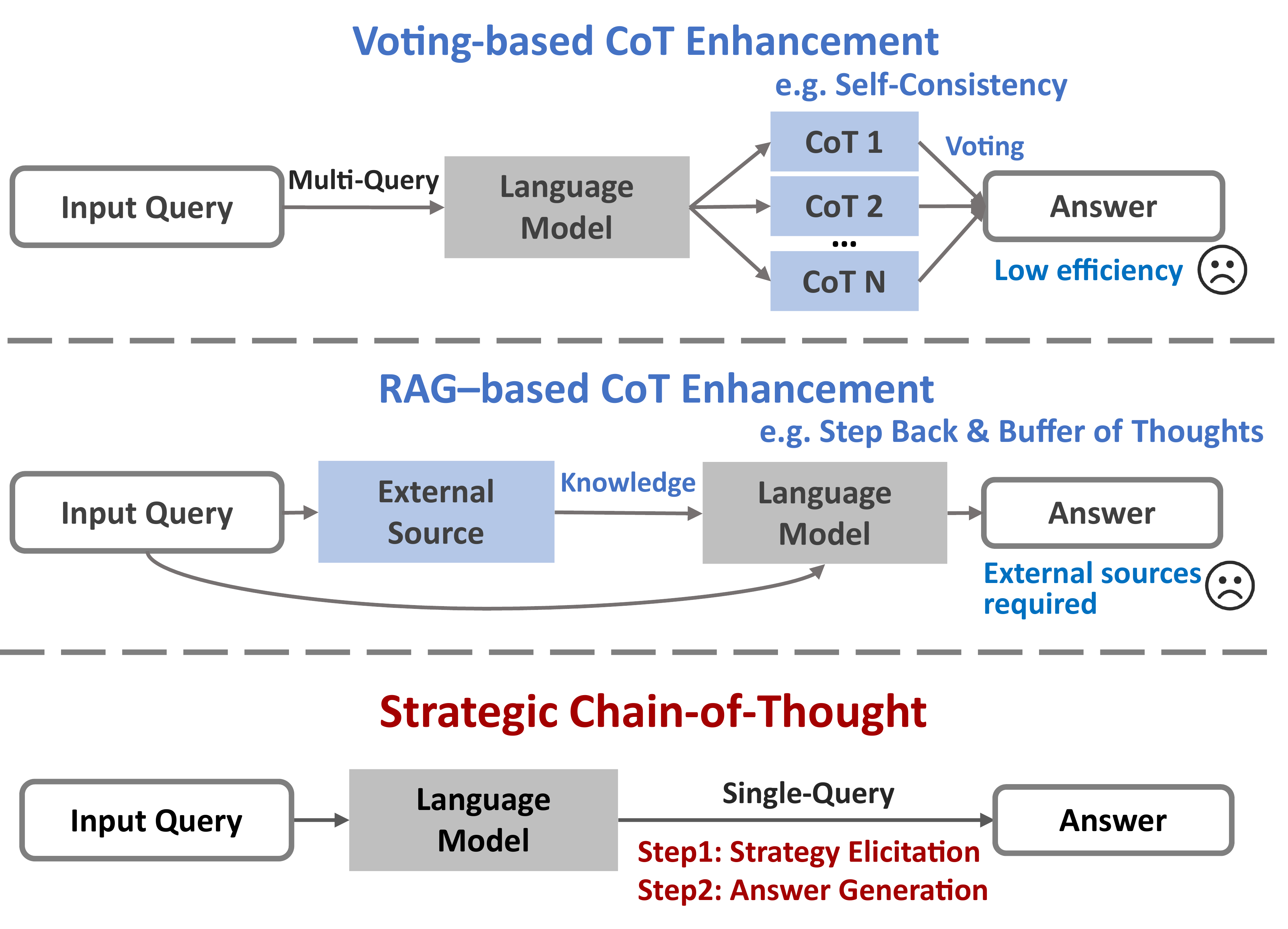}
\caption{Comparison of some popular methods with SCoT: As a single-query method, SCoT is efficient and does not rely on external knowledge sources, distinguishing it from other approaches.}
\label{fig:comrecentandscot}
\end{figure} 

This variability in error probability, influenced by the undetermined strategies used to generate CoT paths, can undermine the reliability of the CoT approach in critical applications where precise and reliable reasoning is essential. Therefore, further refinement and improvement of the CoT methodology are necessary to enhance its performance in complex reasoning scenarios, drawing on insights from both artificial intelligence and cognitive science.

Various methods have been developed to address this challenge by enhancing the quality of CoT paths in LLMs, as illustrated in Figure \ref{fig:comrecentandscot}.
Among these methods, voting-based approaches enhance reasoning accuracy by generating diverse reasoning paths and then voting on the most reliable and correct answer~\cite{wang2022self, zhang2023multimodal}. Retrieval-Augmented Generation (RAG)-based approaches introduce external sources to access additional knowledge through multi-step prompting strategies~\cite{lewis2021retrievalaugmentedgenerationknowledgeintensivenlp, yang2024buffer, zheng2023take}. 
These approaches improve the reasoning process by systematically incorporating and aligning external information before arriving at the final result. Additionally, \citeauthor{suzgun2024meta}~\shortcite{suzgun2024meta} have integrated various prompt enhancement algorithms, dynamically selecting the optimal one to produce the most accurate results during actual operation. 

These approaches do help mitigate the variability in path quality; however, they often come with significant resource demands. For instance, methods like Self-Consistency~\cite{wang2022self} may require up to 40 queries, while techniques such as BoT~\cite{yang2024buffer} involve multi-stage queries. Additionally, some approaches may necessitate the integration of external knowledge sources to achieve optimal performance, which places high demands on expert resources.

To tackle this challenge, we propose a novel approach called Strategic Chain-of-Thought (SCoT). SCoT is designed to improve the quality of CoT path generation for reasoning tasks by incorporating strategic knowledge. The method involves a two-step process within a single prompt. First, it explores and identifies various problem-solving strategies, eliciting the most effective one as the guiding strategic knowledge. Subsequently, this strategic knowledge directs the model in generating high-quality CoT paths and producing accurate final answers, ensuring a more effective reasoning process. 
We further extend the SCoT framework by adapting it to a few-shot method. In this approach, strategic knowledge is used to automatically select the most relevant demonstrations. These examples can be employed within both the few-shot and SCoT frameworks to further enhance reasoning capability.
SCoT enhances the model's reasoning capabilities without the need for multi-query approaches or additional knowledge sources. By eliminating the requirement for multiple queries and external knowledge integration, SCoT reduces computational overhead and operational costs, making it a more practical and resource-efficient solution.

The concept of strategic knowledge in our approach is also inspired by the recent Re-TASK framework~\cite{wang2024retaskrevisitingllmtasks}, which revisits LLM tasks from the perspectives of capability, skill, and knowledge. While Re-TASK enhances LLM capabilities through knowledge injection and skill adaptation via capability items, SCoT takes a different approach by eliciting knowledge rather than relying on explicit knowledge injection. Furthermore, the demonstrations based on strategic knowledge in SCoT are analogous to the capability items in Re-TASK.

We conducted experiments across eight reasoning datasets spanning five distinct domains: mathematical reasoning, commonsense reasoning, physical reasoning, spatial reasoning, and multi-hop reasoning. The results revealed substantial improvements across various models, including a 21.05\% increase in accuracy on the GSM8K dataset and a 24.13\% increase on the Tracking\_Objects dataset with the Llama3-8b model. These results validate the effectiveness of the SCoT approach.

The contributions of this work are summarized as follows:
\begin{itemize}
    \item We introduce a two-stage methodology that integrates strategic knowledge, guiding the LLM to generate high-quality CoT paths by first developing a problem-solving strategy and then producing the final answer.
    \item We propose a method that leverages strategic knowledge to select and match relevant demonstrations, enabling the precise pairing of high-quality CoT examples.
    \item Our experimental results validate the effectiveness of SCoT, demonstrating promising outcomes in reasoning tasks across multiple domains.
\end{itemize}

\section{Related Work}
%Recently, 
\subsection{Strategic Diversity in Problem Solving}
In the realm of problem-solving, there is rarely a one-size-fits-all approach. The complexity of each problem often necessitate a variety of strategies to reach an effective solution. In the fields of education and cognitive science, the phenomenon of using multiple approaches to solve problems is quite common~\cite{swell, rusczyk2003art}. 
Similarly, researchers have found that LLMs might generate diverse solution paths for one question, where the problem-solving strategies and answers of these methods might vary significantly~\cite{wang2024chain, wang2022self}. 

\subsection{Enhancement of CoT Path} 
Current methods for enhancing the quality of model-generated content are diverse and sophisticated.

Some approaches utilize a voting-based mechanism. For example, \citeauthor{wang2022self}~\shortcite{wang2022self} introduced the Self-Consistency method, which improves reasoning accuracy by first generating more than 20 CoT paths and then voting for the most consistent answer.
Other methods incorporate external sources. \citeauthor{zheng2023take}~\shortcite{zheng2023take} introduced Step Back, which prompts models to generate an abstract of the question to capture deeper logical structures, thereby enhancing retrieval-augmented generation (RAG) capabilities. Similarly, \citeauthor{yang2024buffer}~\shortcite{yang2024buffer} developed another RAG-based method, Buffer of Thoughts, which uses knowledge extracted from external sources and predefined knowledge categories for each task. These elements are integrated into a predefined task prompt template, enabling the model to generate more accurate answers.
Additionally, some methods introduce external tools to aid problem-solving. \citeauthor{gao2023pal}~\shortcite{gao2023pal} proposed PAL, which leverages large language models to parse problems and generate programs as intermediate reasoning steps, delegating the solution to a runtime environment like a Python interpreter. This neural-symbolic collaboration has demonstrated improved accuracy across various tasks. \citeauthor{suzgun2024meta}~\shortcite{suzgun2024meta} introduced meta-prompting, which integrates existing prompt-based frameworks, enabling dynamic selection of the most effective reasoning strategy. These strategies, with their complex templates and multi-stage prompting, provide models with sophisticated tools for advancing CoT generation in LLMs.

These methods are inherently complex, with some being task-sensitive and others involving multi-turn prompting; however, they have demonstrated substantial efficacy in enhancing the reasoning capabilities of LLMs, thereby advancing the frontiers of CoT generation in machine learning.

\section{Method}
\begin{figure*}[t]
  \centering
   \subfigure[Framework of Zero-shot and Few-shot Strategic Chain-of-Thought. The solid line in the middle represents Zero-shot SCoT, while the dashed line on the right represents Few-shot SCoT.]{
    \label{fig:cot}
    \includegraphics[width=0.72\textwidth]{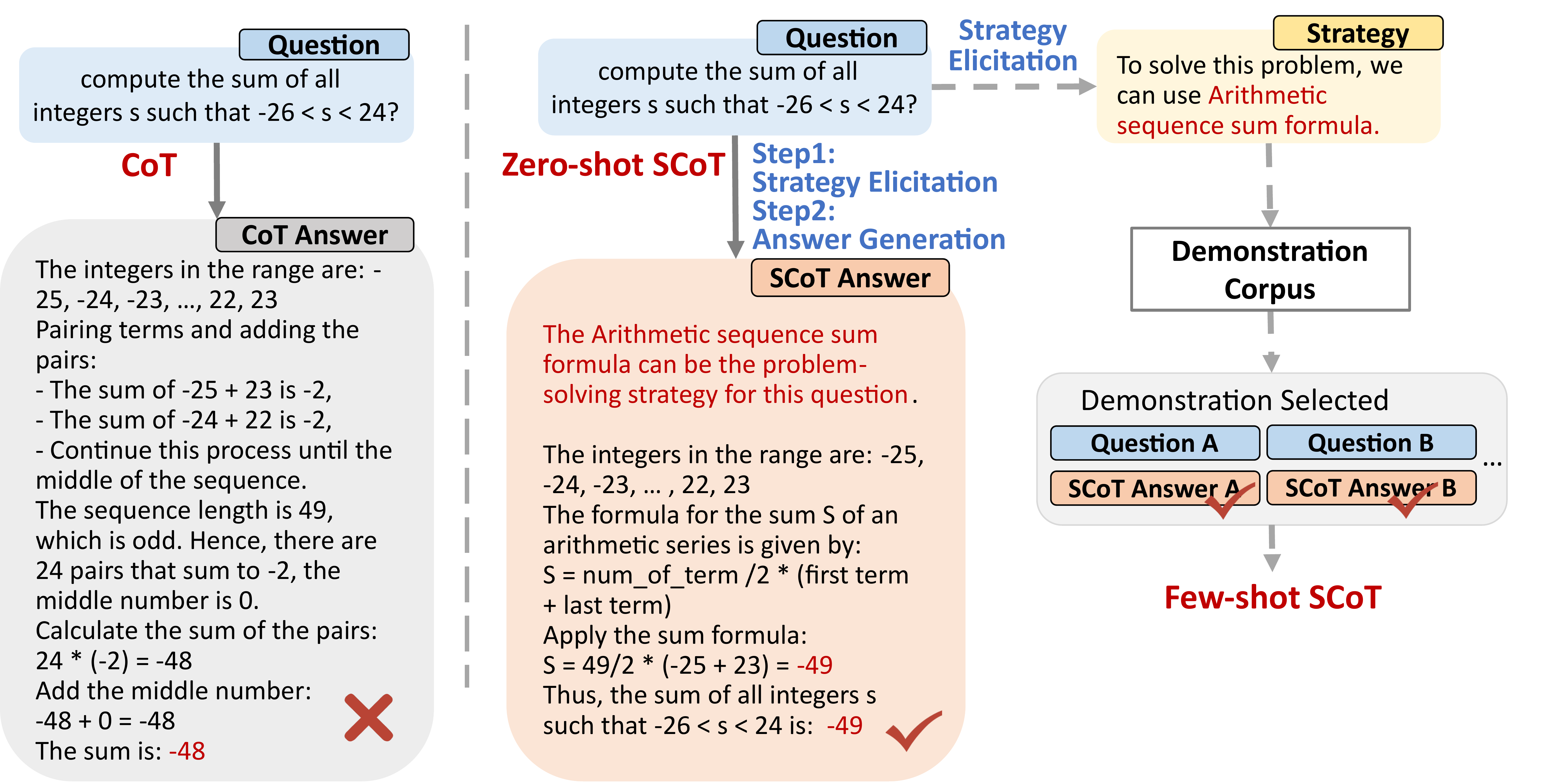}} 
    \hspace{0.1cm}
  \subfigure[Construction of Demonstration Corpus]{
    \label{fig:corpus}
\includegraphics[width=0.235\textwidth]{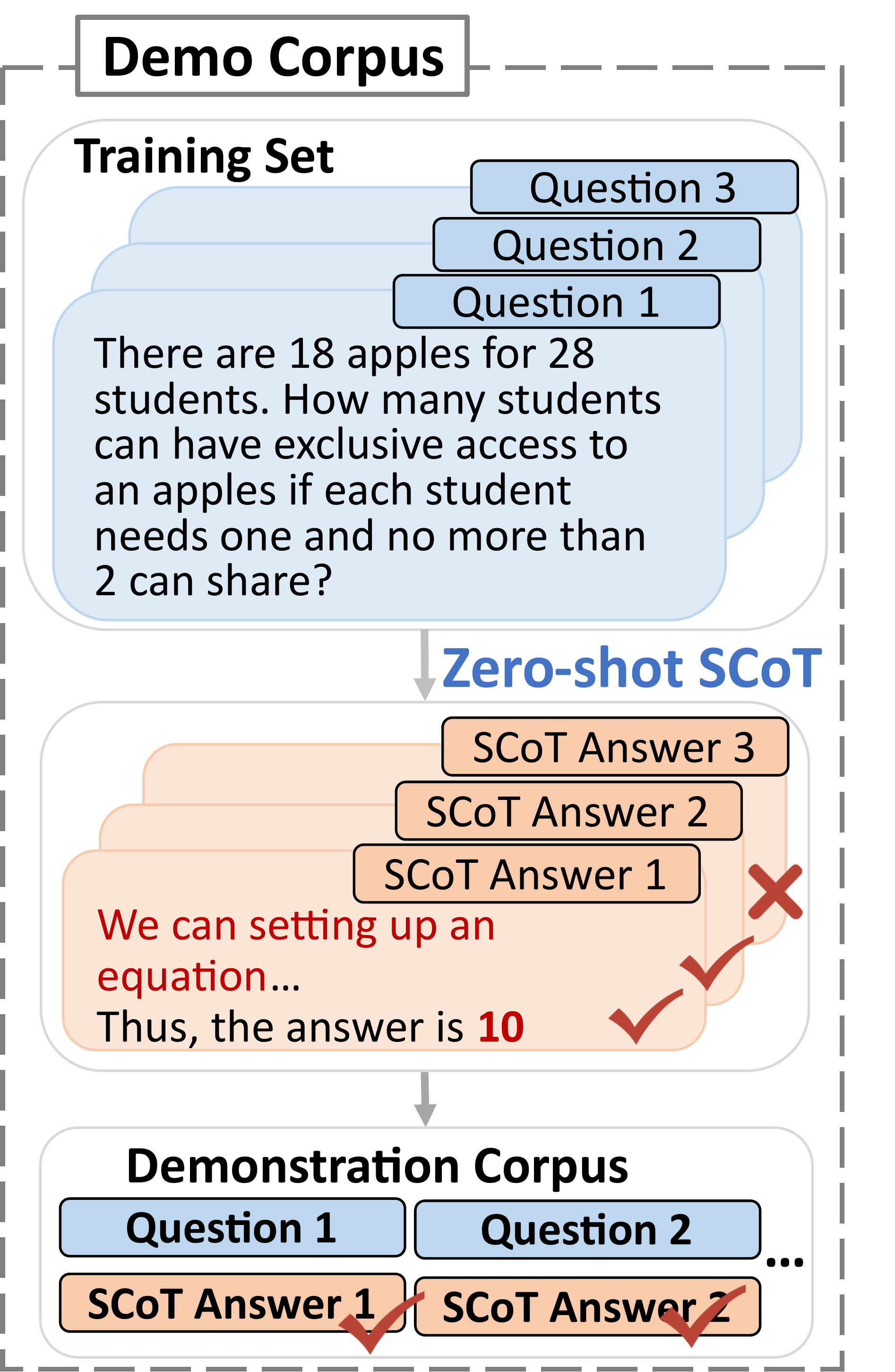}} 
  \caption{Illustration of Zero-shot and Few-shot Strategic SCoT. Few-shot SCoT builds upon Zero-shot SCoT by incorporating selected demonstrations. Details of the Few-shot SCoT approach are omitted due to space limitations.
}
  \label{fig:scot}
\end{figure*}

In this section, we introduce the strategic knowledge, the Strategic Chain-of-Thought (SCoT) method, and its extension through the few-shot approach.

\subsection{Strategic Knowledge}
LLMs tend to produce varied CoT paths for the same problem. However, the quality of these CoT paths can vary significantly~\cite{wang2024chain, wang2022self}.
As shown in the left part of Figure \ref{fig:cot}, when solving the math question "\textit{compute the sum of all integers s such that $-26<s<24$}", one possible approach utilizes term pairing and summing the pairs to generate the final answer. Another possible approach employs the arithmetic series sum formula to compute the final result directly. While both methods are valid for problem-solving, the first approach results in less stable outputs typically due to the complexity of the intermediate steps. In contrast, the second approach, which applies the arithmetic series formula, generally results in better quality and more stable model outputs. The arithmetic series formula is considered strategic knowledge. 

Strategic knowledge (Strategy) refers to a well-defined method or principle that guides reasoning towards a correct and stable solution. 
It involves using structured processes that logically lead to the desired outcome, thereby enhancing the stability of CoT generation and improving the overall quality of the results. 

Specifically, strategic knowledge should adhere to the following principles: 

1. Correct and Comprehensive Problem-Solving Approach: It provides a systematic approach that allows the model to generate accurate answers when it follows the reasoning steps carefully. 

2. Relatively Straightforward Problem-Solving Steps: The steps of the method should not be overly complex, while each step should be sufficiently detailed to ensure accuracy and prevent overly brief outputs that could lead to ambiguity.

\subsection{Strategic Chain-of-Thought}
\label{sec:SCoT}
Building on the concept of strategic knowledge, we propose a prompt-based method to enhance the reasoning quality of LLMs, called Strategic Chain-of-Thought (SCoT).

The SCoT method enables the model to first elicit strategic knowledge before generating an answer, rather than producing an answer directly. Specifically, in a single-query setting, SCoT involves two key steps:

1. \textbf{Elicitation of Strategic Knowledge}: The model identifies and determines one of the most effective and efficient methods for solving the problem, which then serves as the strategic knowledge for the task.

2. \textbf{Application of Strategic Knowledge}: The model subsequently applies the identified strategic knowledge to solve the problem and derive the final answer.

\begin{figure}[t]
  \centering
   \subfigure[SCoT]{
    \label{fig:scotprompt}
    \includegraphics[width=0.22\textwidth]{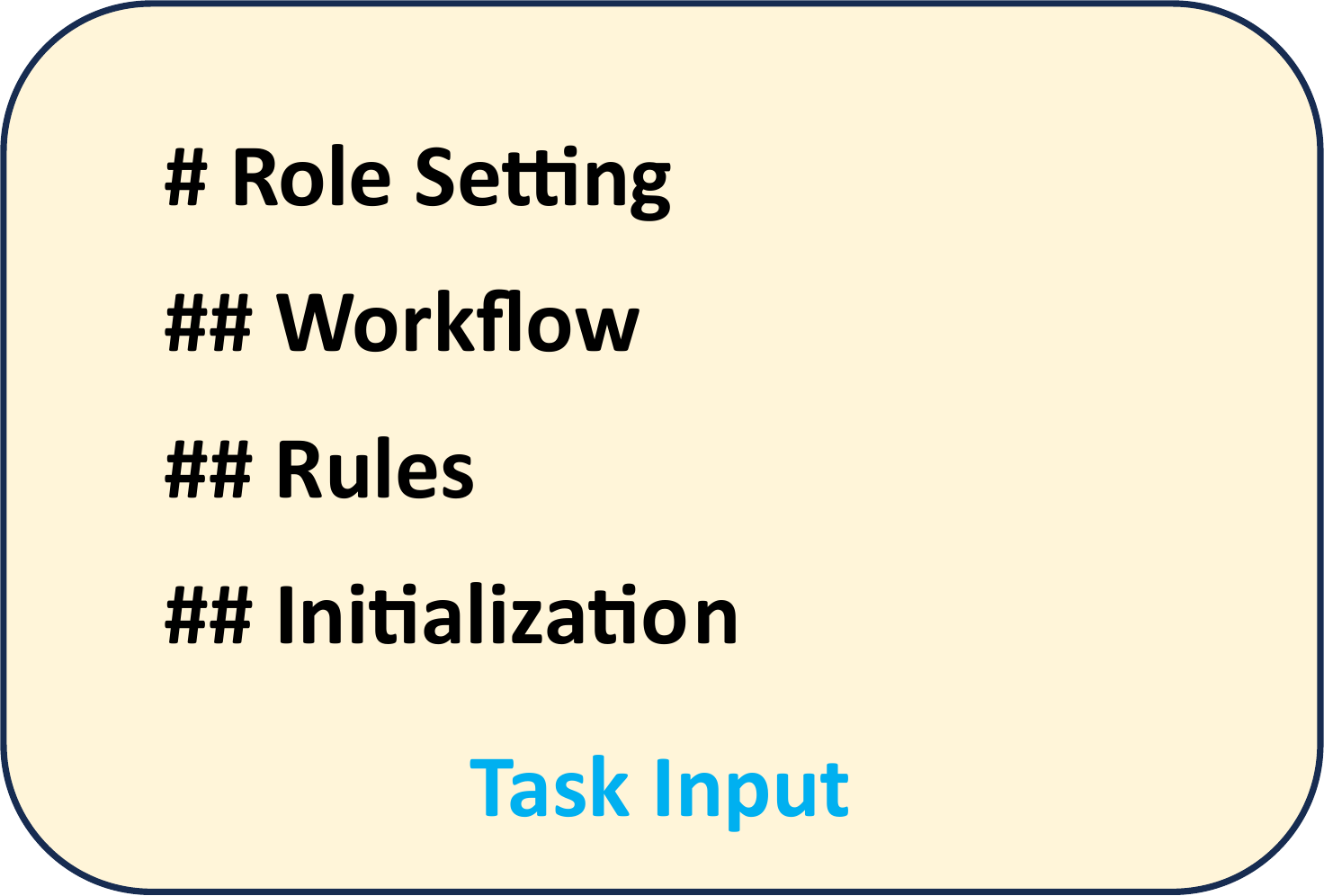}} 
  \subfigure[Few-shot SCoT]{
    \label{fig:fewshotprompt}
    \hspace{0.15cm}\includegraphics[width=0.22\textwidth]{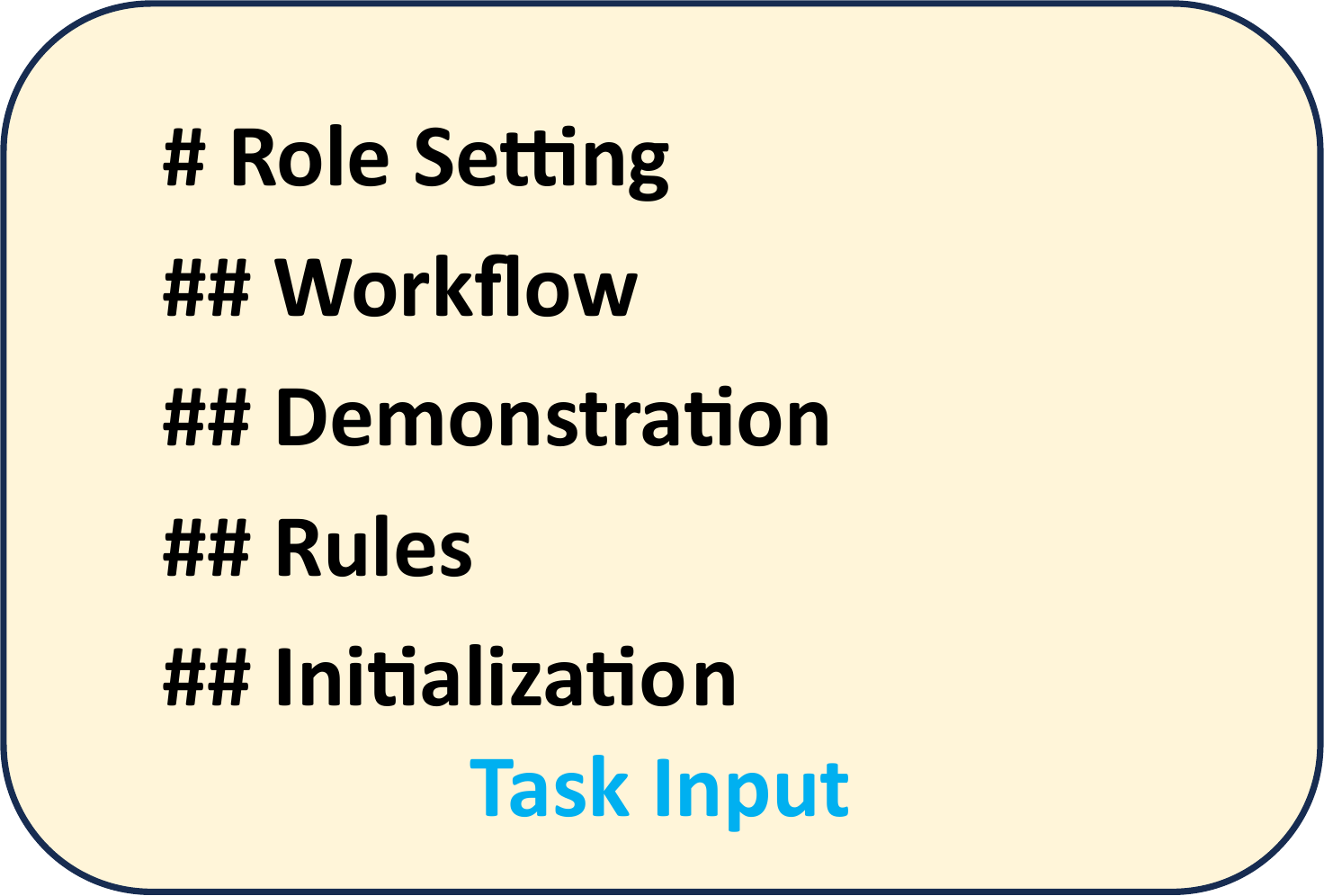}} 
  \caption{Prompt templates for zero-shot and few-shot SCoT
}
  \label{fig:prompt template}
\end{figure}

\begin{figure}[ht]
\centering
\includegraphics[width=0.48\textwidth]{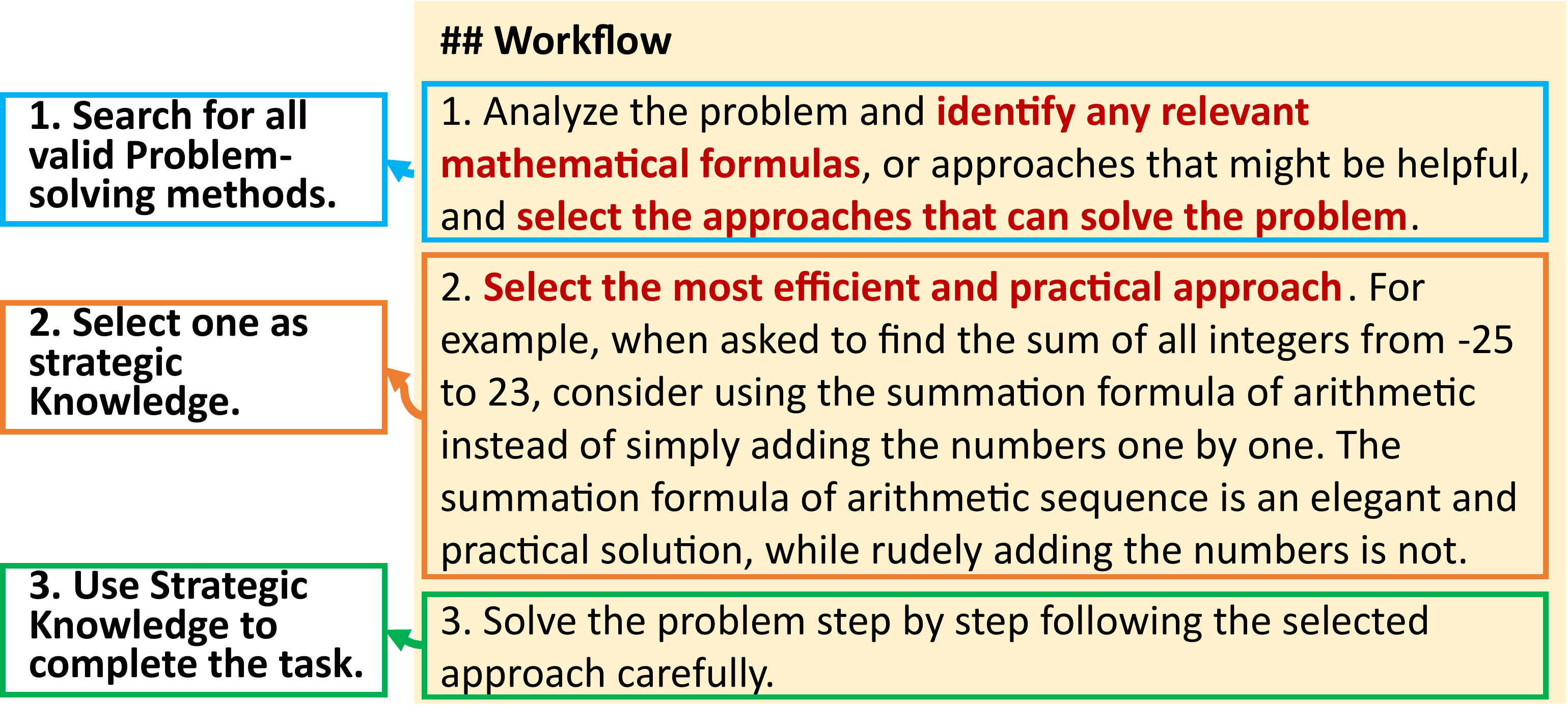}
\caption{Example of a Workflow in a Math Task Prompt}
\label{fig:workflow}
\end{figure} 

Figure \ref{fig:scotprompt} illustrates a prompt template utilizing the SCoT approach.
Our prompt consists of five components: Role, Workflow, Rule, Initialization, and Task Input. 
The prompt incorporates a structured workflow comprising three steps integrated into a single prompt. 
The first two steps are designed to identify and elicit strategic knowledge for solving the problem, while the third step focuses on applying the strategy to generate the answer, as shown in Figure \ref{fig:workflow}.

We demonstrate that the rules for strategic knowledge identification vary across different domains. In mathematics, strategic knowledge favors generating elegant and efficient solutions, such as using the arithmetic series formula to sum sequences. In physics, it involves selecting the most relevant and straightforward formulas or processes, such as applying 
$F=ma$ to calculate force. For multi-hop reasoning, strategic knowledge focuses on determining the appropriate granularity for problem decomposition and recalling pertinent information. Similarly, in other domains, the model first develops an overarching method or workflow before systematically applying it to solve problems, such as optimizing complex systems through algorithms and heuristics.

\subsection{Few-shot Strategic Chain-of-Thought}
We refine the SCoT method into a few-shot version by leveraging the strategy to select demonstrations. Our approach is structured into two stages: constructing a strategy-based demonstration corpus and performing model inference.

\textbf{Stage 1}: Strategic Knowledge-Based Demonstration Corpus Construction.

This stage involves the following two steps, as shown in Figure \ref{fig:corpus}:

1. SCoT Answer Generation: We apply the zero-shot SCoT method to the training set to generate a corresponding SCoT answer for each question in the dataset.

2. Demonstration Corpus Construction: The generated answers are compared with the ground truth. Only those accurate question-SCoT answer pairs are retained. This step assumes that the strategic knowledge used in these problems is both correct and relevant. The validated question-SCoT answer pairs are then compiled into a demonstration corpus based on strategic knowledge.

\textbf{Stage 2}: Model Inference.

This stage involves the following three steps in a two-query process, as shown in the right of Figure \ref{fig:cot}:

1. Strategic Knowledge Generation: The LLM generates strategic knowledge relative to the problem, focusing on understanding the problem rather than producing the final answer.

2. Demonstration Matching: The generated strategic knowledge is used to search the demonstration corpus created in Stage 1. The system identifies and matches the most relevant demonstrations with the SCoT answers from the most similar examples.

3. Few-shot Inference: The selected demonstrations are integrated as few-shot examples into the input prompt (Figure \ref{fig:fewshotprompt}). This integration guides the model to generate the final prediction based on the provided examples.

\section{Experimental Setup}
In this section, we introduce the detailed experimental setup for validation of SCoT, including the datasets used for testing, the models covered, and the baselines employed.
\subsection{Datasets and Tasks}
To validate the effectiveness of the SCoT method, we collect a range of reasoning-related datasets covering domains including mathematics and physical reasoning, commonsense and multi-hop reasoning, and spatial reasoning:

1. Mathematics and Physical Reasoning: We assess the models using datasets such as MathQA~\cite{amini2019mathqainterpretablemathword}, AQuA~\cite{ling2017programinductionrationalegeneration}, GSM8K~\cite{cobbe2021trainingverifierssolvemath}, and MMLU-high-school-math~\cite{hendrycks2021measuringmassivemultitasklanguage} for mathematical reasoning tasks. These datasets feature a range of mathematical problems with varying levels of difficulty, demanding strong mathematical reasoning abilities. Additionally, we evaluated the models on ARC\_Challenge~\cite{clark2018thinksolvedquestionanswering} for physical reasoning, \textit{i.e.}, a popular dataset that presents significant challenges in this domain.

2. Commonsense and Multi-hop Reasoning: We evaluate the models on CommonsenseQA (CSQA)~\cite{talmor-etal-2019-commonsenseqa} for commonsense reasoning tasks and StrategyQA (SQA)~\cite{geva2021didaristotleuselaptop} for multi-hop reasoning tasks. These datasets are well-regarded in their respective domains and offer a substantial level of difficulty.

3. Spatial Reasoning: We also evaluate the models using the Tracking\_Object (Object)~\cite{srivastava2023beyond} dataset, which represents a less common but highly intriguing type of reasoning task.

In the few-shot version of SCoT, we conduct experiments exclusively on the MathQA, AQuA, GSM8K, and ARC datasets. This selection is due to the requirement that the dataset must have a sufficiently large training set with gold answers for constructing the demonstration corpus in the first step. Only these four datasets meet this criterion.

\subsection{Models}
To verify the effectiveness of the SCoT method, we utilize the following LLMs: 
the Llama3 series~\cite{dubey2024llama3herdmodels} (including Llama3-8B, Llama3-70B, Llama3.1-8B, and Llama3.1-70B); the Llama2 series~\cite{touvron2023llama2openfoundation} (including Llama2-7B, Llama2-13B, and Llama2-70B); Mistral-7B~\cite{jiang2023mistral7b}; the Qwen2 series~\cite{qwen2} (including Qwen2-7B and Qwen2-72B); and ChatGLM4-9B~\cite{glm2024chatglm}. ChatGLM4-9B is chat-oriented and other models are instruction-tuned.

\input{Tables/All_results_2models}
\input{Tables/All_results_3datasets}
\subsection{Baselines}
We use zero-shot prompts~\cite{kojima2022large}, Self-Consistency~\cite{wang2022self} and Step Back~\cite{zheng2023take} as baselines. We only conducted experiments on 5 datasets using Step Back because Step Back is not well-suited for other datasets. BoT~\cite{yang2024buffer} is not chosen because its template has not been available, making it impossible to reproduce.

We select the accuracy as the metric for the performance, which is calculated by the average results of three independent inferences on each model. The experimental parameter settings are provided in the appendix. 

\section{Experimental Results}
In this section, we empirically evaluate the effectiveness of the Strategic Chain-of-Thought (SCoT) approach. To verify SCoT's efficacy across all datasets, we test it using two open-source models, Llama3-8B and Mistral-7B.
To further validate SCoT's effectiveness across different models, we select one dataset from each of the three reasoning task categories and conduct tests on all 7 models. We also examine the impact of model size, perform ablation studies on SCoT components, conduct case studies, and analyze experimental efficiency to understand the factors influencing the effectiveness of SCoT.

\subsection{Results across all Datasets}
\label{sec:alldata}
The experimental results across all datasets using two models are presented in Table \ref{tab:Mainresult1}. 
Notably, in zero-shot settings, SCoT outperforms the CoT approach in most tasks, with particularly significant improvements observed on the GSM8K dataset, where accuracy increases from 52.11\% to 73.16\% after incorporating strategic knowledge. Additionally, SCoT achieves a 24.13\% improvement on the Tracking\_Object dataset. However, the Llama3-8B model exhibits a 2.6\% decrease in performance on the ARC dataset. In general, the Llama3-8B model shows an average improvement of 6.92\% on all datasets, while the Mistral-7B model demonstrates an average improvement of 3.81\% on comparable datasets. Compared to Step Back and Self-Consistency, SCoT also performs better than these two methods except for the result of Self-Consistency with Llama3-8B model on the ARC dataset. Nevertheless, our SCoT still achieves comparable results to it.
Notably, SCoT shows substantial gains in commonsense reasoning tasks compared with other methods.

Furthermore, we extend the SCoT framework to support few-shot settings by automatically matching demonstrations, resulting in even stronger performance. The $\text{SCoT 1-shot}^-$, as shown in Table \ref{tab:Mainresult1}, refers to CoT prompting with demonstrations matched through strategic knowledge. Compared to CoT 0-shot\footnote{We do not present the accuracy of CoT 1-shot separately as it was comparable to CoT 0-shot in our experiments.}, $\text{SCoT 1-shot}^-$, which uses strategy-matched demonstrations, shows significant performance improvements across most datasets, highlighting the effectiveness of the matched demonstrations. The $\text{SCoT 1-shot}$, which combines both strategic knowledge and strategy-matched demonstrations, achieves the best results overall.

\subsection{Results across all Models}

The experimental results for all models on the three datasets are shown in Table \ref{tab:mainresult2}. The experiments demonstrate that SCoT can enhance performance across most models. In particular, with the exception of the Llama3.1-8B model, where the addition of SCoT results in a slight decrease in accuracy on the MMLU task, other models exhibit accuracy improvements ranging from 1.11\% to 24.13\% across the three datasets. Note that the CoT 0-shot has achieved 100\% accuracy with Llama3.1-70B model on Tracking\_Object dataset, and SCoT 0-shot maintains this performance.

\label{sec:modelscale}
\begin{figure}[htb]
\centering
\includegraphics[width=0.48\textwidth]{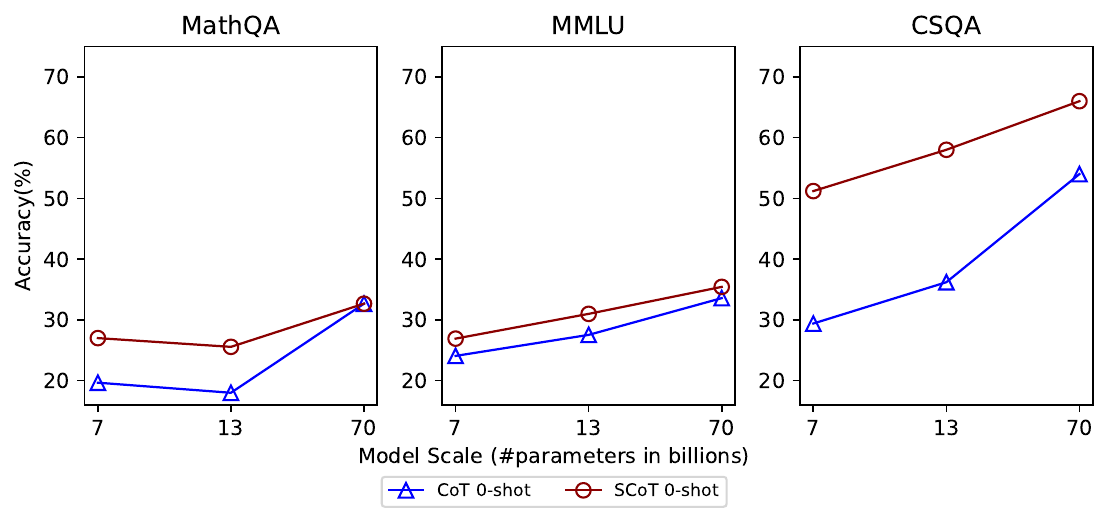}
\caption{Accuracy(\%) across three datasets using different scales of models in Llama2 series}
\label{fig:ModelScale}
\end{figure} 
\subsection{Model Scale}
Here we investigate the impact of model size on the effectiveness of SCoT. Experiments on the Llama2 model series with three different sizes are conducted, and the results are shown in Figure \ref{fig:ModelScale}. It demonstrates that SCoT can lead to accuracy improvements across all sizes of the Llama2 models. However, a general trend emerges that performance improvement decreases marginally with model size. Furthermore, manual inspection of the model outputs reveals that larger models are more likely to generate CoT path containing strategic knowledge in 0-shot settings.

\subsection{Ablation Study}
\input{Tables/AblationStudy}
We explore the effects of various components within the prompt (such as role, workflow, structure, and the quantity of demonstrations) on accuracy. The experimental results are illustrated in Table \ref{tab:abstudy}. Building on the CoT 0-shot approach, we observed that adding roles, incorporating workflows, and formatting prompts in markdown progressively increased accuracy. We also explored the impact of the number of demonstrations on accuracy within the few-shot SCoT framework. Experimental results indicate that as the number of demonstrations increases, the performance of SCoT either slightly improves or remains unchanged.

\subsection{Case Study}
\begin{figure*}[t]
\centering
\includegraphics[width=0.96\textwidth]{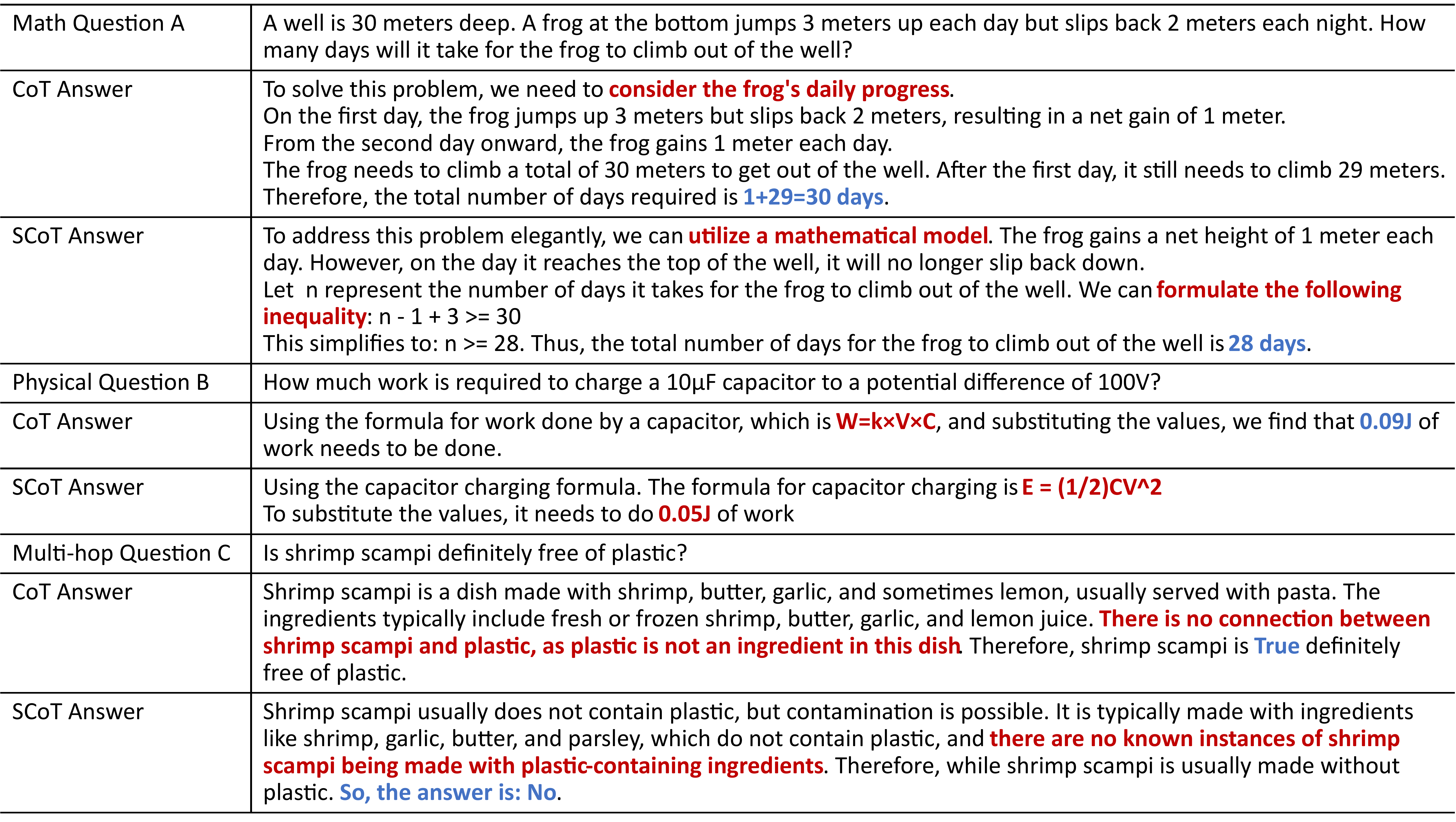}
\caption{Comparison of the paths generated by CoT and SCoT in different domains.}
\label{fig:casestudy}
\end{figure*} 

We conduct a detailed case study focusing on the validity of the strategic knowledge elicited from the model. Figure \ref{fig:casestudy} shows several typical cases.

In the domain of mathematics, we observe that the SCoT output tends to favor solving problems using inequalities rather than directly analyzing the problem to reach an answer. For the instance of frog jumping calculation in the Figure \ref{fig:casestudy}, an incorrect solution may miscalculate the final jump's impact. While generating a strategy ensures accurate calculations by considering all constraints and systematically solving the problem.

In the field of physics, we find that the model's CoT output could be misled by specific phrases in the task input (\textit{e.g.}, "capacitor"), leading to the selection of an incorrect formula. In contrast, the SCoT approach successfully elicited the correct formula. Similarly, in multi-hop reasoning tasks, CoT output often focuses on details, resulting in incomplete subsequent logical reasoning, whereas SCoT generates answers by considering the overall context.

\subsection{Efficiency Analysis}
\input{Tables/Efficient}
Due to SCoT's mechanism of generating strategy before solving problems in one query, it is more efficient than multi-query methods. However, compared to single-query methods, the output token length might be longer, potentially decreasing efficiency. To investigate this, we measure the output token lengths for the AQuA, GSM8K, and Tracking\_Object datasets using both CoT 0-shot and SCoT 0-shot methods. The results are shown in Table \ref{tab:efficient}.

The results indicate that the token length output by the Mistral-7B model on the GSM8K dataset decreases with the SCoT method. This reduction may be due to the model's tendency to repetitively generate a specific answer span up to the inference length limit on the GSM8K dataset in CoT 0-shot, leading to a decline in accuracy. SCoT mitigates this issue. Besides, the length of SCoT varies from 1.03 to 1.8 times that of CoT, averaging around 1.5 times. This shows that while our method is somewhat slower than CoT, the efficiency remains manageable.

\section{Discussions}
\subsection{Automatic SCoT}
To demonstrate that our experimental results are not influenced by human-crafted prompts but rather due to the concept of SCoT,
we conduct a preliminary test to evaluate whether the SCoT prompt templates can be automatically generated. We provide the SCoT concept to Qwen2-72B to generate the corresponding prompt templates and tested these on the AQuA dataset. The results are presented in Table \ref{tab:auto}. The findings indicate that while the accuracy of prompts automatically generated based on the SCoT concept is lower than that of manually crafted SCoT prompts, it is still superior to 0-shot CoT performance. This suggests that the automatic generation of SCoT-based prompt templates is feasible.
\input{Tables/AutoPrompt}

\section{Conclusion}
In this paper, we introduce the Strategic Chain-of-Thought, a method that enables LLMs to autonomously generate an optimal Chain-of-Thought path. By integrating a structured workflow for eliciting and applying strategic knowledge, SCoT enhances the model's ability to produce a high quality outputs.
We further extend SCoT to a few-shot version by matching demonstrations through strategic knowledge from a predefined strategic knowledge-based corpus. Experimental results demonstrate the effectiveness of both 0-shot SCoT and few-shot SCoT.

Overall, SCoT offers a promising framework for improving the quality of reasoning path in LLMs. Future research will focus on evaluating its effectiveness with more complex problems and exploring further applications.

\bibliography{aaai25}

\clearpage

\appendix

\section{Details of Experiments}
\subsection{Models Details}
\input{appendixTables/Modelsources}
This experiment involves ten models, nine of which are public (Llama3-8B, Llama2-7B, Mistral-7B, Llama3.1-8B, Qwen2-7B, ChatGLM4-9B, Llama3-70B, Llama3.1-70B, Llama2-70B, and Qwen2-72B), while one model is private. The sources and licenses for all public models are detailed in Table \ref{modelsources}.

\subsection{Datasets Details}
\input{appendixTables/DataSources}
This experiment involves eight datasets: MathQA, AQuA, GSM8K, MMLU, ARC, StrategyQA, CommonsenseQA, and Tracking\_Object. All datasets used in this study are publicly available, with their sources and licenses detailed in Table \ref{datasources}.

MathQA, AQuA, MMLU, ARC, StrategyQA, CommonsenseQA, and Tracking\_Object consist of multiple-choice questions. To determine correctness, we compare the predicted choice with the gold (correct) choice. For GSM8K, the answers are numerical text spans; we assess correctness by checking if the predicted answer exactly matches the gold answer.

\subsection{Other Details}
For all experiments, except those involving Self-Consistency, the temperature is set to 0, and the top\_p parameter is set to 1. For Self-Consistency, following the settings from the original paper~\cite{wang2022self}, the temperature is adjusted to 0.5, and top\_p is set to 0.5.

We utilize vllm~\cite{kwon2023efficient} as the inference framework for all deployments.

\section{Results}
\subsection{All Results}

Accuracy is used as the evaluation metric. We conducted three independent inference runs for all experiments and calculated the average results. However, due to the high computational cost, we performed only a single inference for Self-Consistency. The accuracy and standard deviation results are presented in Table \ref{tab:Mainresult1full} and Table \ref{tab:mainresult2full}.

\input{appendixTables/all_results_2model_full}
\input{appendixTables/all_results_3datasets_full}

\subsection{Case Study}
\begin{figure*}[t]
\centering
\includegraphics[width=0.85\textwidth]{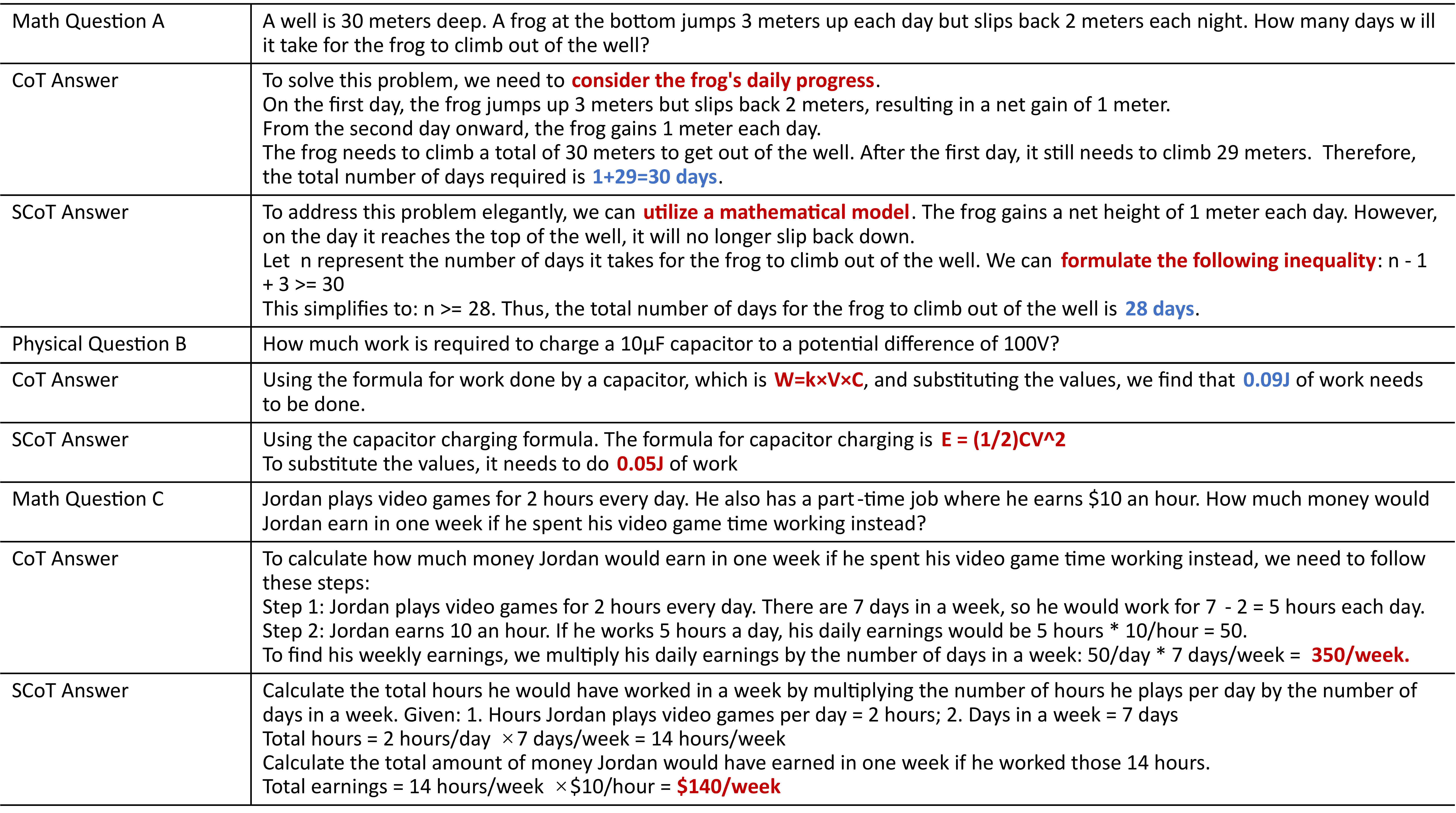}
\caption{Comparison of the paths generated by CoT and SCoT in different domains.}
\label{appendfig:casestudy1}
\end{figure*}

\begin{figure*}[t]
\centering
\includegraphics[width=0.85\textwidth]{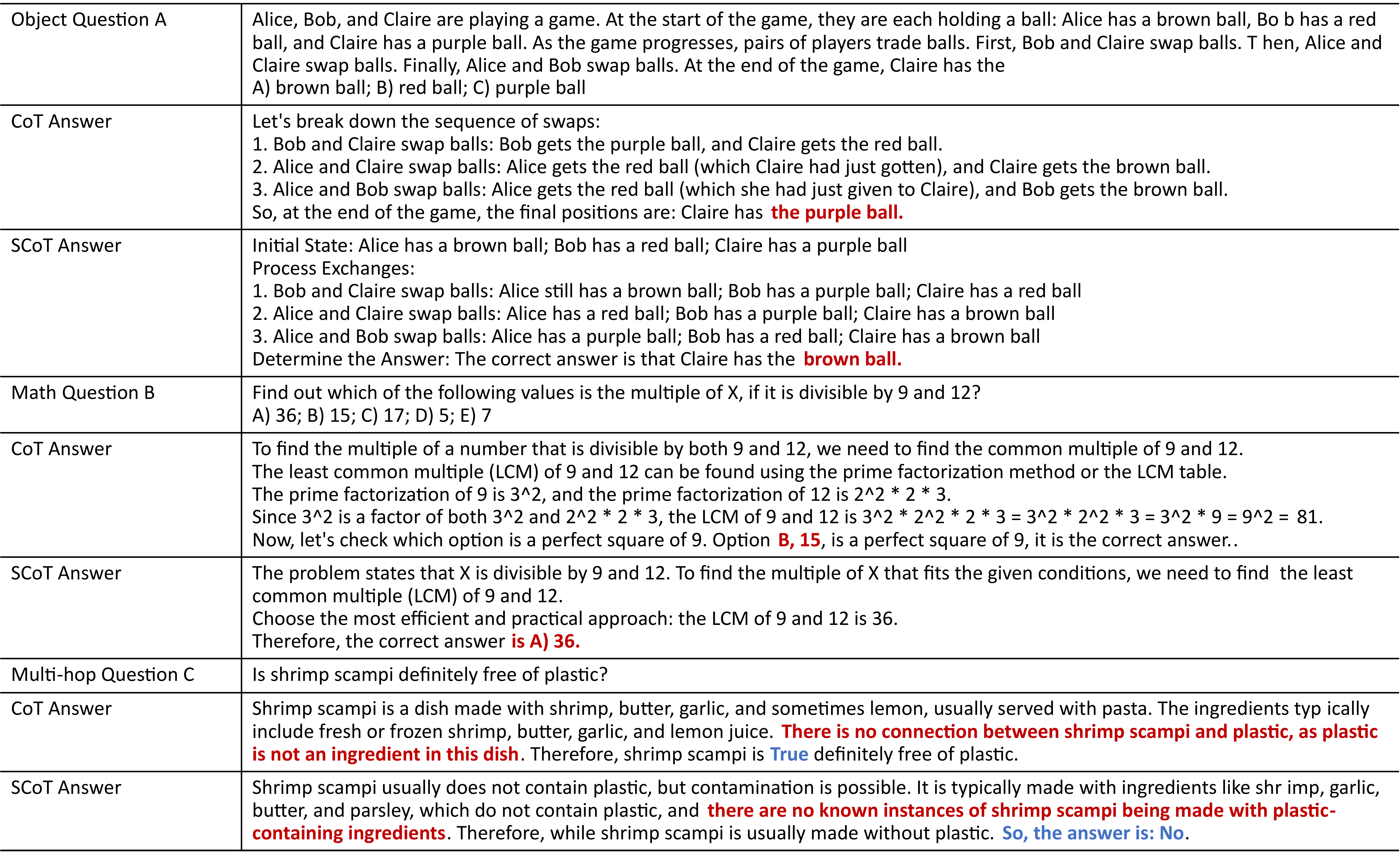}
\caption{Comparison of the paths generated by CoT and SCoT in different domains.}
\label{appendfig:casestudy2}
\end{figure*}

We conducted a detailed case study to assess the validity of the strategic knowledge elicited from the model. Figures \ref{appendfig:casestudy1} and \ref{appendfig:casestudy2}  present several representative cases spanning math reasoning, physical reasoning, commonsense reasoning, multi-hop reasoning, and spatial reasoning.

\section{Experimental Prompts}
\begin{figure*}[t]
\centering
\includegraphics[width=0.65\textwidth]{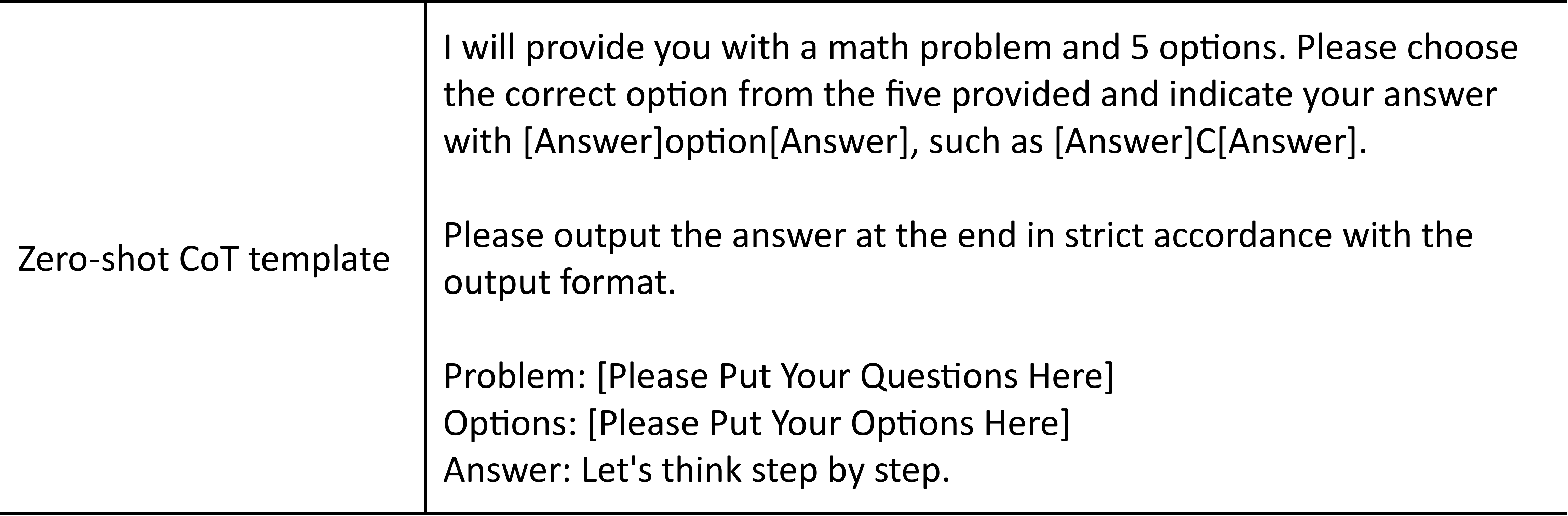}
\caption{An example of prompting for standard zero-shot CoT}
\label{appendixfig:cot}
\end{figure*}

\begin{figure*}[t]
\centering
\includegraphics[width=0.8\textwidth]{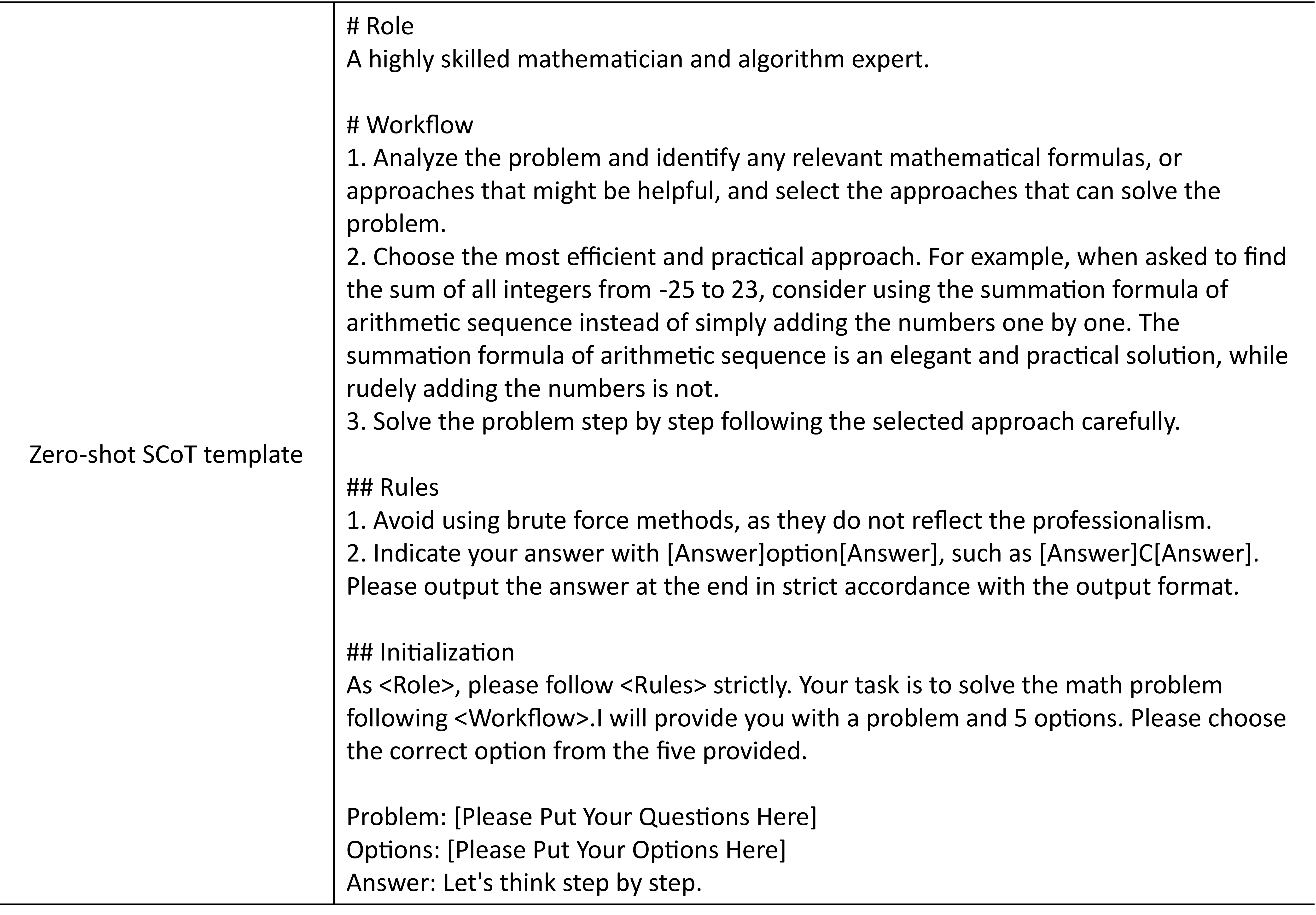}
\caption{An example of prompting for standard Strategic Chain-of-Thought in math reasoning tasks}
\label{appendixfig:scot-math}
\end{figure*}

\begin{figure*}[t]
\centering
\includegraphics[width=0.8\textwidth]{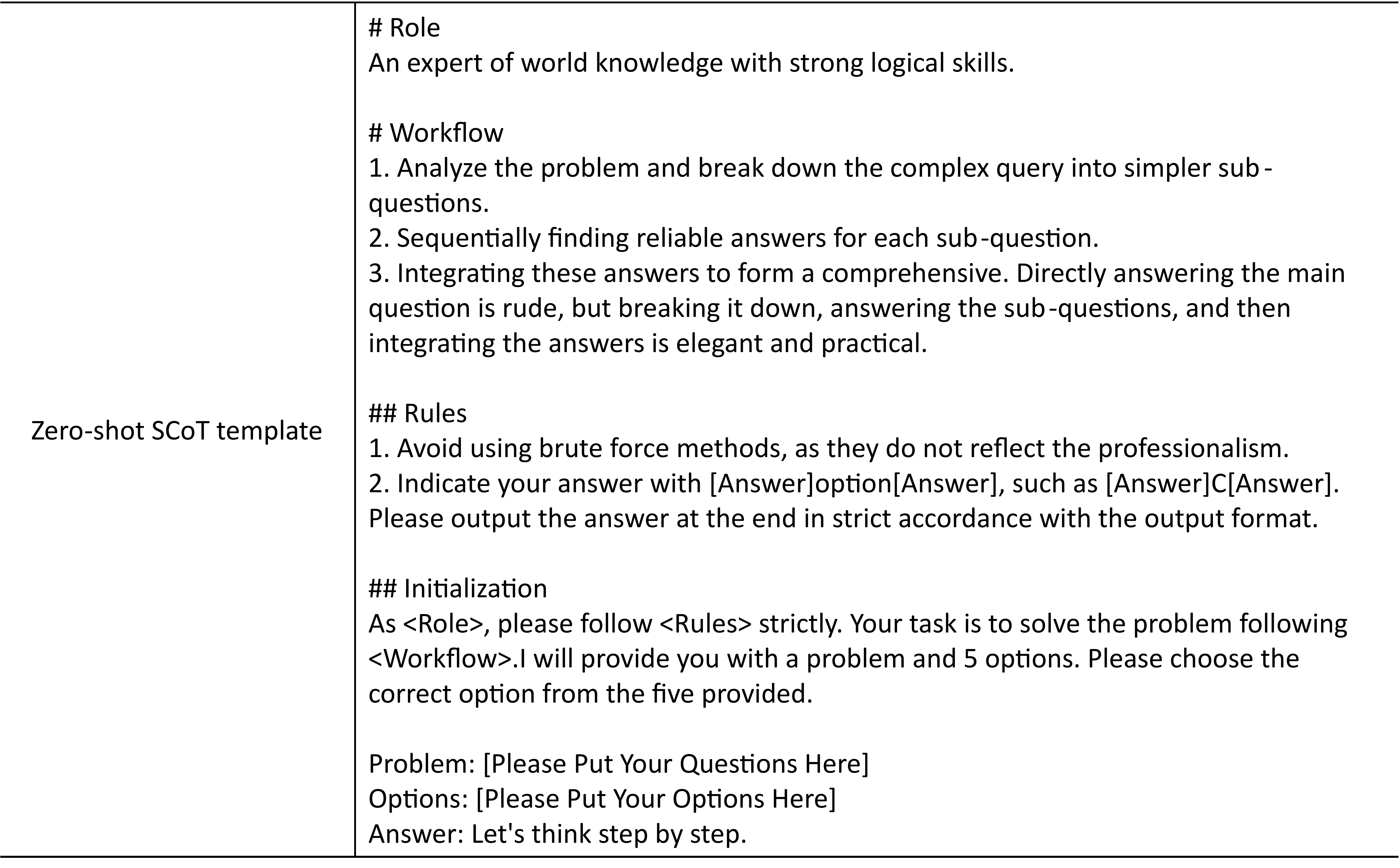}
\caption{An example of prompting for standard Strategic Chain-of-Thought in multi-hop reasoning tasks}
\label{appendixfig:scot-multihop}
\end{figure*}

\begin{figure*}[t]
\centering
\includegraphics[width=0.8\textwidth]{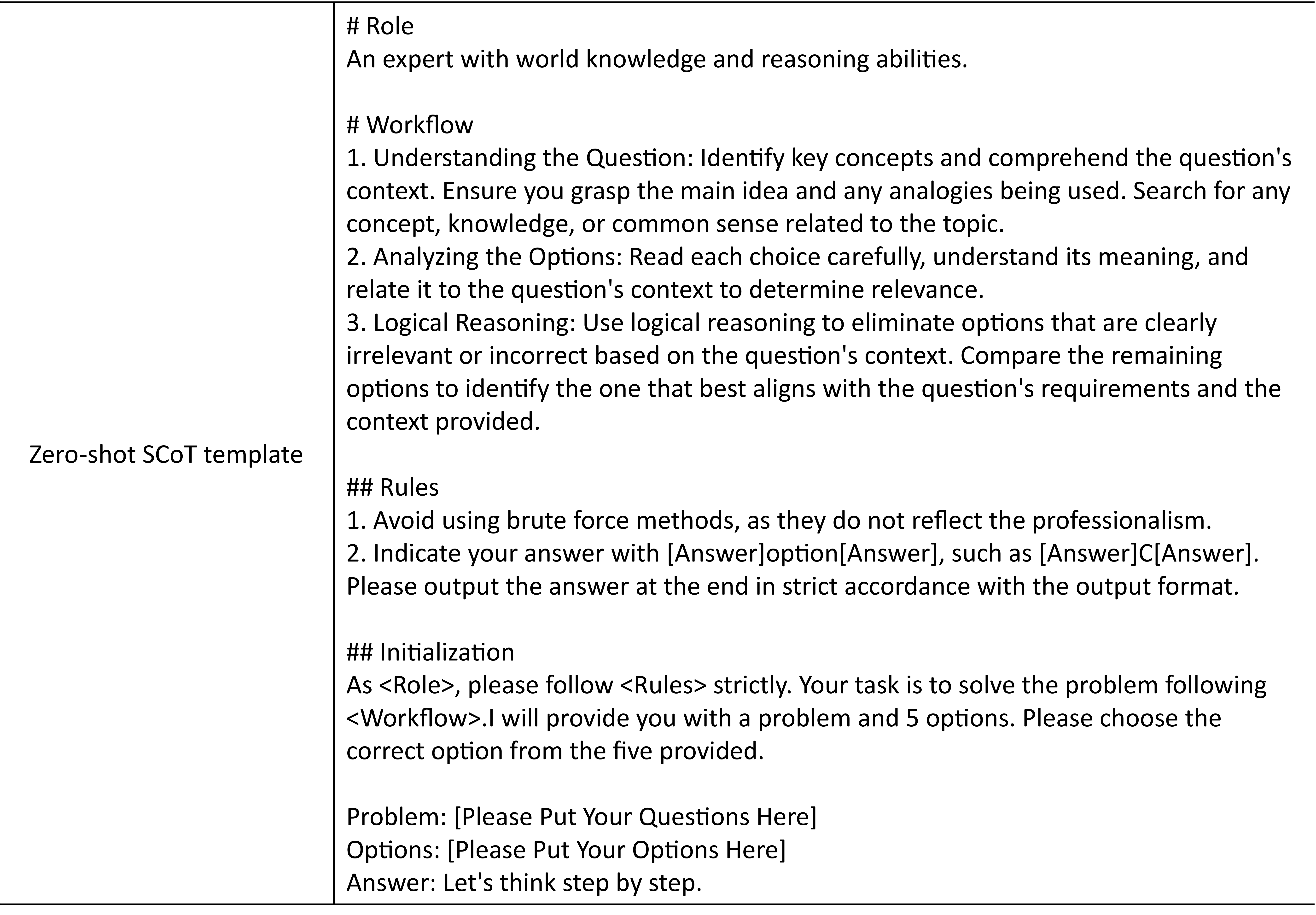}
\caption{An example of prompting for standard Strategic Chain-of-Thought in commonsense reasoning tasks}
\label{appendixfig:scot-commonsense}
\end{figure*}

\begin{figure*}[t]
\centering
\includegraphics[width=0.8\textwidth]{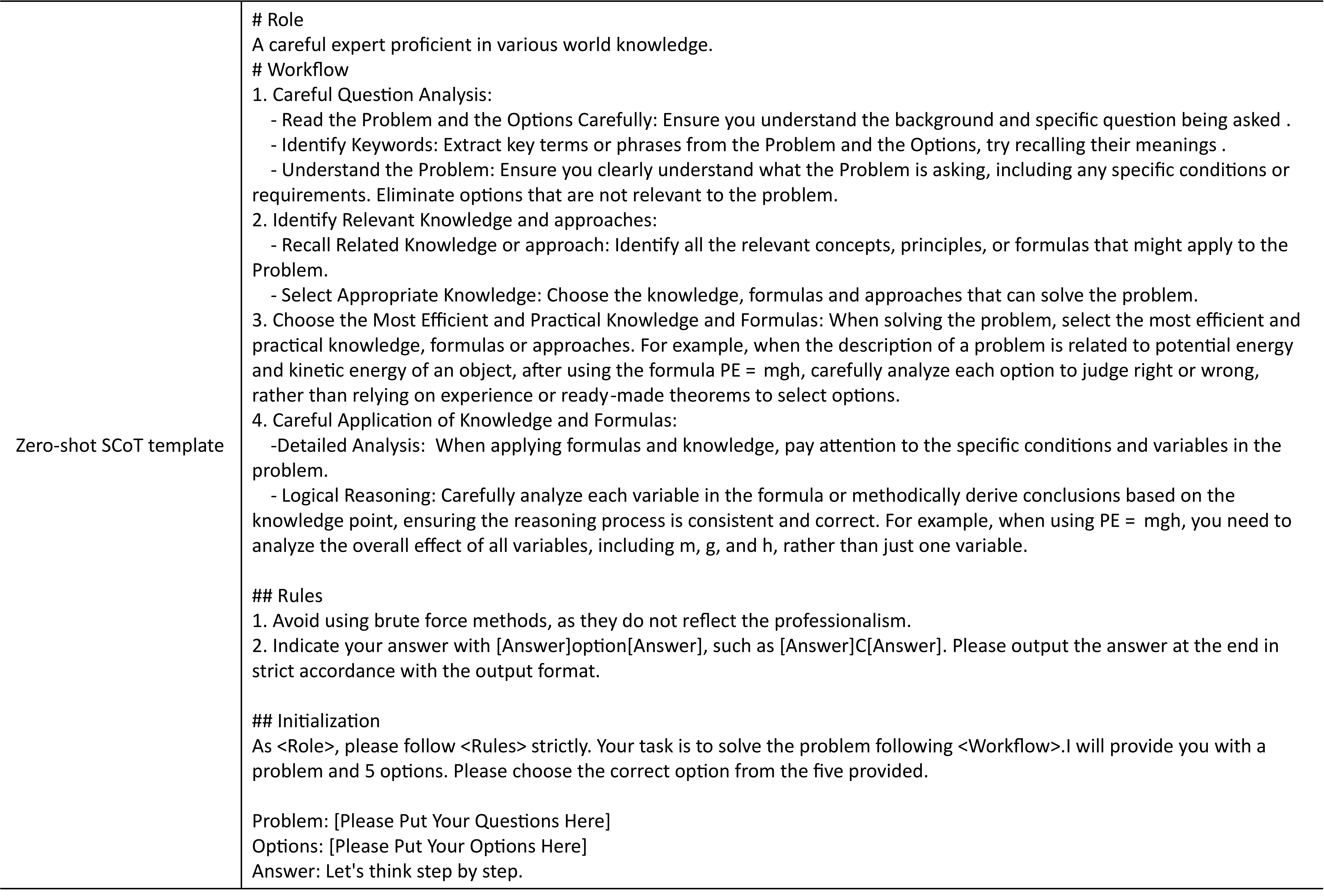}
\caption{An example of prompting for standard Strategic Chain-of-Thought in physical reasoning tasks}
\label{appendixfig:scot-physical}
\end{figure*}

\begin{figure*}[t]
\centering
\includegraphics[width=0.8\textwidth]{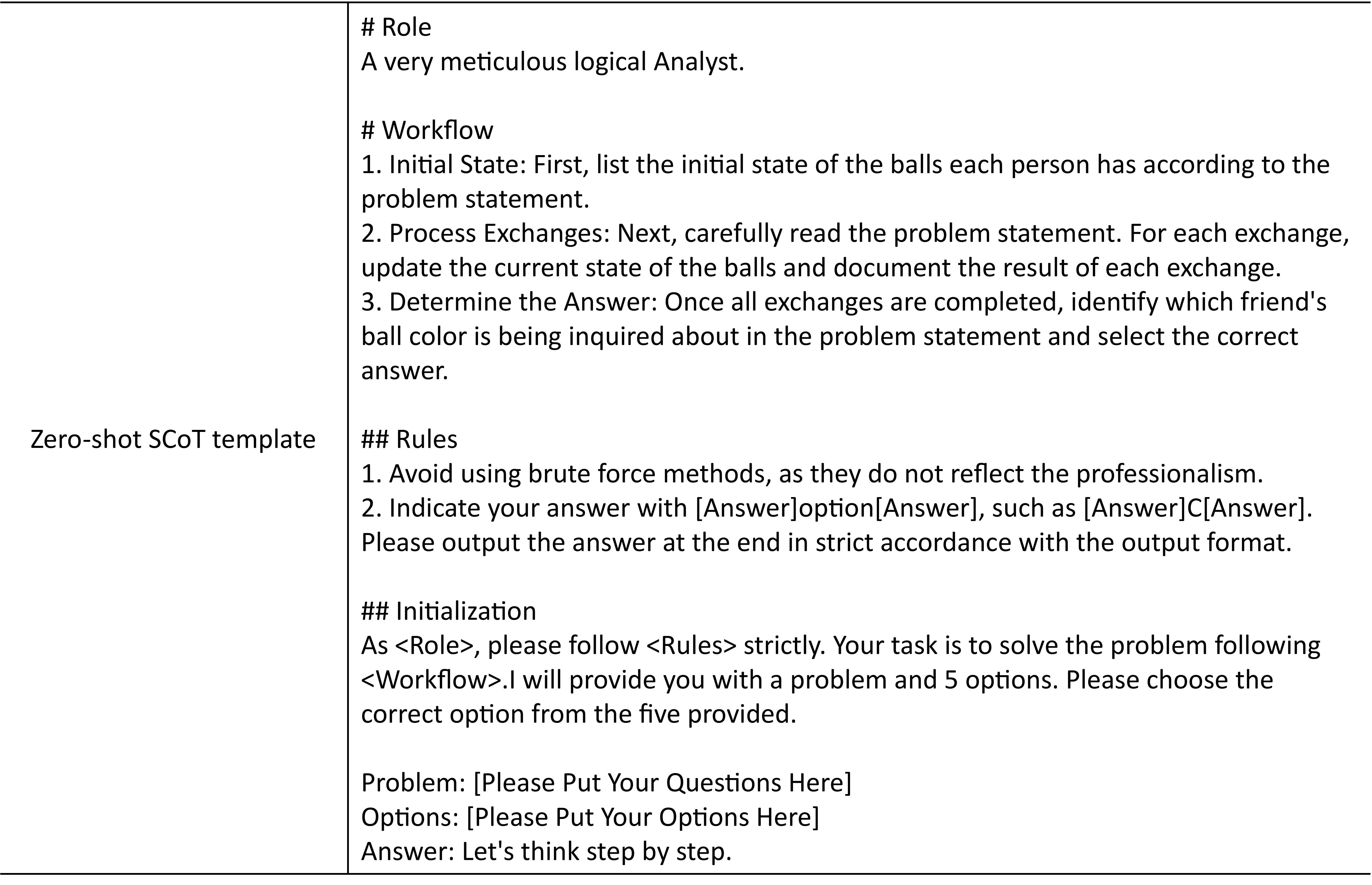}
\caption{An example of prompting for standard Strategic Chain-of-Thought in spatial reasoning tasks}
\label{appendixfig:scot-spatial}
\end{figure*}

\begin{figure*}[t]
\centering
\includegraphics[width=0.8\textwidth]{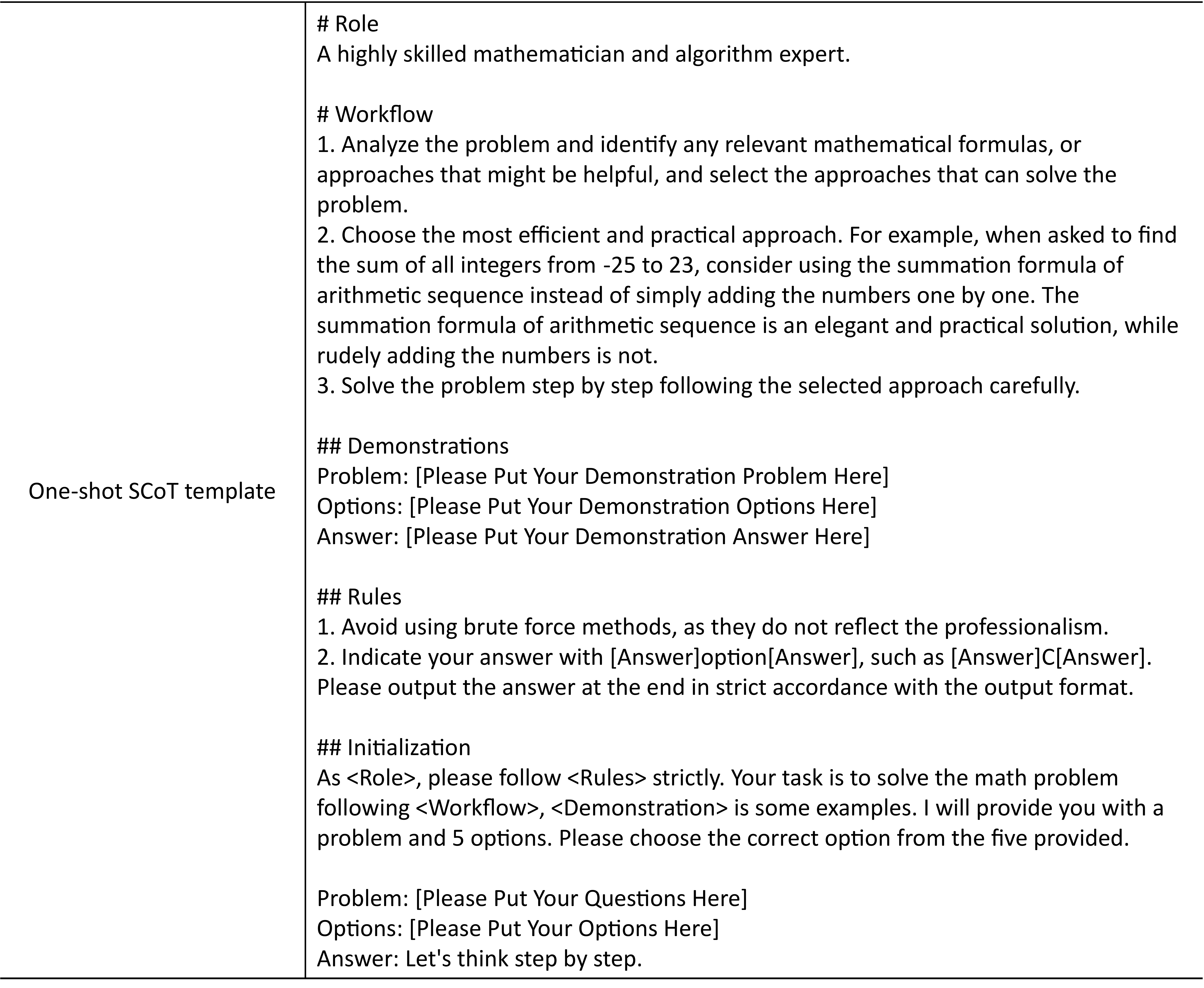}
\caption{An example of prompting for one-shot Strategic Chain-of-Thought}
\label{appendixfig:1scot}
\end{figure*}

\begin{figure*}[t]
\centering
\includegraphics[width=0.6\textwidth]{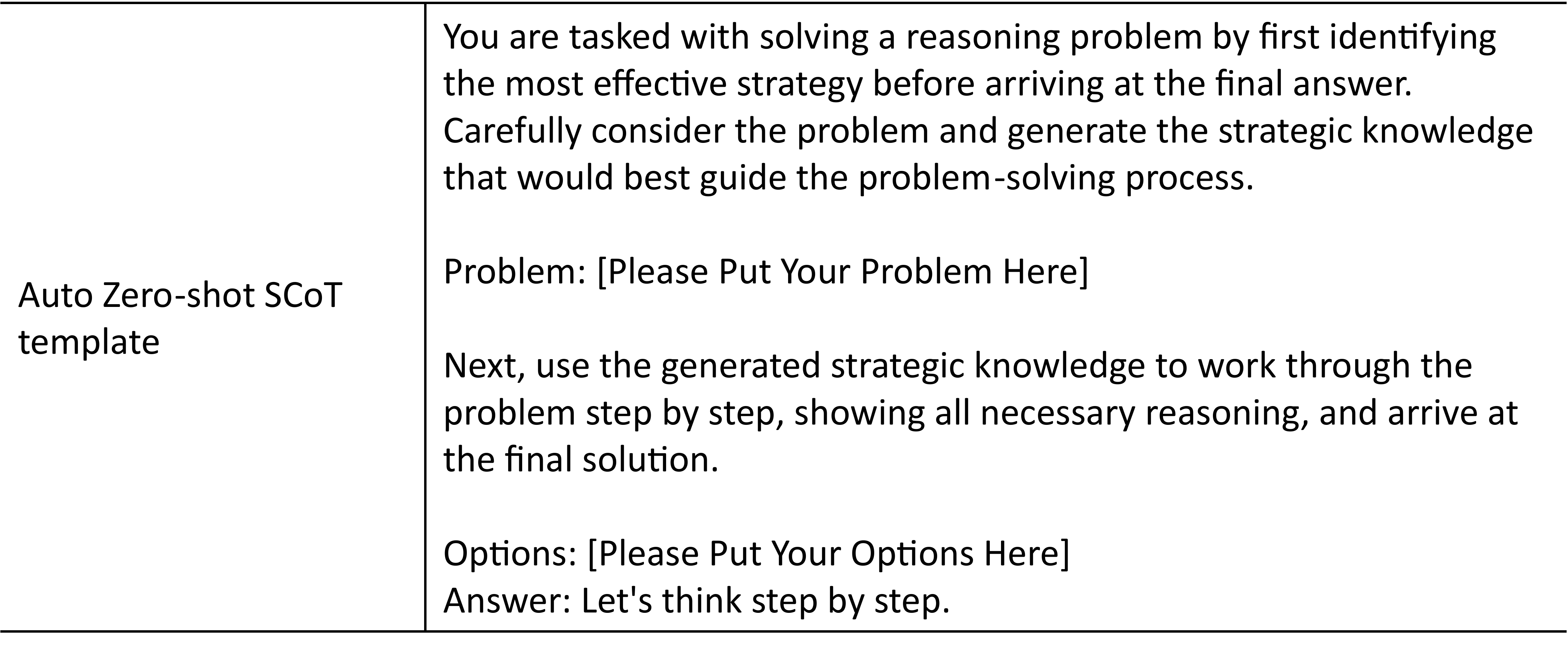}
\caption{An example of prompting for automatic Strategic Chain-of-Thought}
\label{appendixfig:autoscot}
\end{figure*}

The prompt for standard zero-shot Chain-of-Thought is shown in Figure \ref{appendixfig:cot}. Prompts for zero-shot Strategic Chain-of-Thought are displayed in Figure \ref{appendixfig:scot-math} (for math reasoning), Figure \ref{appendixfig:scot-multihop} (for multi-hop reasoning), Figure \ref{appendixfig:scot-physical} (for physical reasoning), Figure \ref{appendixfig:scot-commonsense} (for commonsense reasoning) and Figure \ref{appendixfig:scot-spatial} (for spatial reasoning). Prompts for one-shot Strategic Chain-of-Thought are shown in Figure \ref{appendixfig:1scot}. Finally, the prompts for automated Strategic Chain-of-Thought are shown in Figure \ref{appendixfig:autoscot}. The automated SCOT prompts were generated using LLMs by given the idea of SCoT.

\end{document}

%% file: Tables/All_results_2models.tex
\begin{table*}[ht]
\centering
\renewcommand{\arraystretch}{1.25}
\small
\begin{tabularx}{\textwidth}{p{1.4cm}p{1.6cm}XXXXXXXX}
\hline
Model             & Method & MathQA & AQuA & GSM8K & MMLU & ARC & SQA & CSQA & Object \\ \cline{1-10}
\multirow{7}{*}{Llama3-8B} & CoT 0-shot      
& 56.33  & 49.61 & 52.11 & 46.67 & 80.60 & 64.60 & 71.13 & 44.27        \\

    & Self-Con  & \textbf{57.00}  & \textbf{51.90} & 48.48 &  \textbf{49.52}         & \textbf{81.00}  & 66.00 & 72.06 & 54.00         \\
    
     & Step Back        & 56.33  & 50.39 &   -- & 47.78 &  75.80  &  64.64 & --  & --        \\
     
     & SCoT 0-shot & \textbf{56.67} & \textbf{51.85} & \textbf{73.16} & \textbf{50.00} & 78.02 & \textbf{68.56} & \textbf{74.00} & \textbf{68.40}        \\ \cline{2-10}
    
    & $\text{SCoT 1-shot}^-$ &  56.33   &  50.87  &    74.91  & -- & 73.40  & --  & -- & --  \\
    
    & SCoT 1-shot      &  \textbf{57.67} & \textbf{55.12} & \textbf{76.57} & -- & \textbf{80.60} & -- &  -- &  --    \\\hline

\multirow{7}{*}{Mistral-7B} & CoT 0-shot & 30.00 & 29.13 & 36.26 & 29.75 & 67.20 & 56.22 & 61.80 & 21.40         \\
    
& Self-Con  & \textbf{31.42}  &  32.87  & 34.50 & 31.88 & \textbf{68.78} & 53.50 &  62.69  & \textbf{24.50}        \\
& Step Back  & 31.43 & 32.87  & --  &  31.85 & 68.00   &  56.72        & --  & --        \\
    & SCoT 0-shot & \textbf{30.44} & \textbf{33.60} & \textbf{38.97} & \textbf{32.35} & \textbf{72.20} & \textbf{61.89} & \textbf{68.00} & \textbf{24.75}      \\ \cline{2-10}
                        
 & $\text{SCoT 1-shot}^-$  & 34.33 & 31.50 & 45.57 & -- & 67.40 & --  & -- & --        \\
& SCoT 1-shot  & \textbf{37.00} & \textbf{35.04}  & \textbf{47.38} & -- & \textbf{73.20} & -- & -- & --    \\ \hline  
\end{tabularx}
\caption{Accuracy (\%) using Llama2-8B and Mistral-7B across all datasets. $\text{SCoT 1-shot}^-$ refers to the results obtained using the standard few-shot CoT template but with demonstrations matched by strategy.}
\label{tab:Mainresult1}
\end{table*}

%% file: Tables/All_results_3datasets.tex
\begin{table*}[ht]
\centering
\small
\renewcommand{\arraystretch}{1.25}
\begin{tabularx}{\textwidth}{cp{1.65cm}lllllllll}
\hline

Dataset &
  Method &
  Llama3-8B &
  Mistral-7b &
  Chatglm4-9B &
  Qwen2-7B &
  Qwen2-70B &
  Llama3.1-8B &
  Llama3.1-70B \\ \hline
  
\multirow{2}{*}{\begin{tabular}[c]{@{}c@{}}MMLU\\ Math\end{tabular}} &
  CoT 0-shot &
  46.67 & 29.75 & 66.67 & \textbf{71.97}  & 84.20  & \textbf{59.63} & \textbf{85.19}   \\
 & SCoT 0-shot &
  \textbf{50.00}\textsubscript{+3.33}    & \textbf{32.35}\textsubscript{+2.59} & \textbf{68.15}\textsubscript{+1.48} & \textbf{71.85} & \textbf{85.93}\textsubscript{+1.73} & 56.42 & \textbf{85.19}  \\ \hline
                     
\multirow{2}{*}{SQA} & CoT 0-shot &
64.60 & 56.22 & 61.80 & \textbf{61.00} & 75.22 & 73.11 & 64.67 \\
                     & SCoT 0-shot      &
 \textbf{68.56}\textsubscript{+3.96} & \textbf{61.89}\textsubscript{+5.67} & \textbf{64.67}\textsubscript{+2.87} & \textbf{61.00} & \textbf{77.67}\textsubscript{+2.45} & \textbf{74.22}\textsubscript{+1.11} & \textbf{82.33}\textsubscript{+1.33}  \\ \hline
                     
\multirow{2}{*}{Object} &
  CoT 0-shot &
  44.27 & 21.40 & 61.80 & 46.20 & 93.93 & 62.60 & \textbf{100.00}   \\
                     & SCoT 0-shot       & 
  \textbf{68.40}\textsubscript{+24.13} & \textbf{24.67}\textsubscript{+3.27} & \textbf{69.00}\textsubscript{+7.20} & \textbf{47.53}\textsubscript{+1.33} & \textbf{97.47}\textsubscript{+3.54} & \textbf{77.60}\textsubscript{+15.00} & \textbf{100.00}  \\ \hline
\end{tabularx}
\caption{Accuracy(\%) across seven models on MMLU, SQA and Tracking\_Object datasets}
\label{tab:mainresult2}
\end{table*}

%% file: Tables/AblationStudy.tex
\begin{table}[ht]
\centering
\renewcommand{\arraystretch}{1.25}
\begin{tabular}{l|p{1.25cm}p{1.25cm}}
\hline
Method & AQuA & ARC \\ \hline
Mistral-7B* & 29.13\% & 67.20\% \\
Mistral-7B + Role* & 27.95\% & 69.80\% \\
Mistral-7B + Role & 32.28\% & 71.20\% \\
Mistral-7B + WorkFlow* & 33.07\% & 70.40\% \\
Mistral-7B + WorkFlow & 31.89\% & 70.40\% \\ \hline
SCoT 0-shot (Ours) & \textbf{33.60\%} & \textbf{72.20\%} \\
SCoT 1-shot (Ours) & \textbf{35.04\%} & \textbf{73.20\%} \\
SCoT 3-shot (Ours) & \textbf{35.43\%} & \textbf{73.20\%} \\ \hline
\end{tabular}
\caption{Ablation study on SCoT prompt components: * denotes a non-markdown format, while no * indicates a markdown format.}
\label{tab:abstudy}
\end{table}

%% file: Tables/Efficient.tex
\begin{table}[ht]
\centering
\renewcommand{\arraystretch}{1.25}
\begin{tabular}{c|c|c|c}
\hline
{Dataset}       & {Method} & {Llama3-8B} & {Mistral-7B}         \\ \hline

\multirow{2}{*}{AQuA}  & CoT 0-shot      & 361.384            & 270.260                  \\ \cline{2-4}

                       & SCoT 0-shot     & \textbf{370.378}   & \textbf{458.413}     \\ \hline
                       
\multirow{2}{*}{GSM8K} & CoT 0-shot      & 130.532            & \textbf{858.507}   \\ \cline{2-4} 
                       & SCoT 0-shot     & \textbf{206.278}   & 611.848            
                       \\ \hline
\multirow{2}{*}{Object} & CoT 0-shot & 121.460 & 89.654 \\ \cline{2-4} 
                       & SCoT 0-shot     & \textbf{174.888}   & \textbf{162.822}    \\ \hline
\end{tabular}
\caption{Token length comparison for SCoT and CoT 0-shot methods}
\label{tab:efficient}
\end{table}

%% file: Tables/AutoPrompt.tex
\begin{table}[ht]
\centering
\renewcommand{\arraystretch}{1.25}
\begin{tabular}{cc}
\hline
Method  & Accuracy \\ \hline
CoT 0-shot       & 29.13             \\
SCoT 0-shot      & 33.60              \\
\textbf{Auto SCoT}      & \textbf{31.89}
\\ \hline
\end{tabular}
\caption{Accuracy(\%) using automatically generated prompts by LLMs based on the SCoT concept}
\label{tab:auto}
\end{table}

%% file: appendixTables/Modelsources.tex
\begin{table*}[ht]
\centering
\small
\renewcommand{\arraystretch}{1.6}
\begin{tabular}{llc}
\hline
\textbf{Models} & \multicolumn{1}{c}{\textbf{Modelsources}}                    & \textbf{License}            \\ \hline
Llama2-7B-chat &   https://huggingface.co/meta-llama/Llama-2-7b-chat & llama2 license          \\
Llama2-13B     & https://huggingface.co/meta-llama/Llama-2-13b-chat & llama2 license       \\
Llama2-70B     & https://huggingface.co/meta-llama/Llama-2-70b-chat & llama2 license        \\
Llama3-8B                       & \renewcommand{\arraystretch}{1.1}\begin{tabular}[c]{@{}l@{}}https://www.modelscope.cn/models/FlagAlpha/\\ Llama3-Chinese-8B-Instruct/summary\end{tabular} & Apache License 2.0                   \\
Llama3.1-8B     & https://huggingface.co/meta-llama/Meta-Llama-3.1-8B-Instruct & llama3.1 license            \\
Llama3.1-70B                    & https://huggingface.co/meta-llama/Meta-Llama-3.1-70B-Instruct                 & llama3.1 license           \\
Mistral-7B      & https://huggingface.co/mistralai/Mistral-7B-Instruct-v0.2    & Apache License 2.0          \\
Qwen2-7B        & https://huggingface.co/Qwen/Qwen2-7B-Instruct                & Apache License 2.0 \\
\multicolumn{1}{l}{Qwen2-72B}   & https://huggingface.co/Qwen/Qwen2-72B-Instruct                                & Apache License 2.0                   \\
\multicolumn{1}{l}{ChatGLM4-9b} & https://huggingface.co/THUDM/glm-4-9b-chat                                    & \multicolumn{1}{l}{glm-4-9b License}\\ \hline
\end{tabular}
\caption{Models, sources and licenses used in this work}
\label{modelsources}
\end{table*}

%% file: appendixTables/Datasources.tex
\begin{table*}[ht]
\centering
\small
\renewcommand{\arraystretch}{1.6}
\begin{tabular}{lll}
\hline
\textbf{Datasets} & \textbf{Sources}                                 & \textbf{Licenses}    \\ \hline
MathQA            & https://huggingface.co/datasets/datafreak/MathQA & Apache License 2.0   \\
AQuA              & https://github.com/google-deepmind/AQuA          & Apache License 2.0   \\
GSM8K             & https://huggingface.co/datasets/openai/gsm8k     & MIT License          \\
MMLU              & https://huggingface.co/datasets/cais/mmlu        & MIT License          \\
ARC               & https://huggingface.co/datasets/allenai/ai2\_arc & CC-BY-SA-4.0 License \\
StrategyQA    & https://huggingface.co/datasets/ChilleD/StrategyQA/viewer/default/test & MIT License \\
CommonsenseQA & https://huggingface.co/datasets/tau/commonsense\_qa                    & MIT License \\
Object Tracking  &    \renewcommand{\arraystretch}{1.05}\begin{tabular}[c]{@{}l@{}}https://github.com/google/BIG-bench/tree/092b196c1f8f14a54bbc62f24759d43bde46dd3b\\ /bigbench/benchmark\_tasks/tracking\_shuffled\_objects/three\_objects\end{tabular} &  Apache License 2.0
\\ \hline
\end{tabular}
\caption{Datasets, sources and licenses used in this work}
\label{datasources}
\end{table*}

%% file: appendixTables/all_results_2model_full.tex
\begin{table*}[ht]
\centering
\small
\begin{tabularx}{0.98\textwidth}{p{1.4cm}p{1.6cm}XXXXXXXX}
\hline
Model             & Method & MathQA & AQuA & GSM8K & MMLU & ARC & SQA & CSQA & Object \\ \cline{1-10}
\multirow{7}{*}{Llama3-8B} & CoT 0-shot      
& 56.33\textsubscript{±0.000}  & 49.61\textsubscript{±1.790} & 52.11\textsubscript{±0.129} & 46.67\textsubscript{±0.000} & 80.60\textsubscript{±0.000} & 64.60\textsubscript{±0.646} & 71.13\textsubscript{±0.094} & 44.27\textsubscript{±0.736}        \\

& Self-Con  & \textbf{57.00}  & \textbf{51.90} & 48.48 &  \textbf{49.52}        & \textbf{81.00}  & 66.00 & 72.06 & 54.00         \\
    
     & Step Back        & 56.33\textsubscript{±0.272}  & 50.39\textsubscript{±0.000} &   -- & 47.78\textsubscript{±0.000} &  75.80\textsubscript{±0.248}  &  64.64\textsubscript{±0.2722} & --  & --        \\
     
     & SCoT 0-shot & \textbf{56.67}\textsubscript{±0.000} & \textbf{51.85}\textsubscript{±1.299} & \textbf{73.16}\textsubscript{±0.163} & \textbf{50.00}\textsubscript{±0.000} & 78.02\textsubscript{±0.000} & \textbf{68.56}\textsubscript{±0.566} & \textbf{74.00}\textsubscript{±0.000} & \textbf{68.40}\textsubscript{±0.000}        \\ \cline{2-10}
    
    & $\text{SCoT 1-shot}^-$ &  56.33\textsubscript{±0.000}   &  50.87\textsubscript{±2.140}  &    74.91\textsubscript{±0.000}  & -- & 73.40\textsubscript{±0.000}  & --  & -- & --  \\
    
    & SCoT 1-shot      &  \textbf{57.67}\textsubscript{±0.000} & \textbf{55.12}\textsubscript{±0.000} & \textbf{76.57}\textsubscript{±0.000} & -- & \textbf{80.60}\textsubscript{±0.000} & -- &  -- &  --    \\\hline

\multirow{7}{*}{Mistral-7B} & CoT 0-shot & 30.00\textsubscript{±0.000} & 29.13\textsubscript{±1.245} & 36.26\textsubscript{±1.854} & 29.75\textsubscript{±0.924} & 67.20\textsubscript{±0.356} & 56.22\textsubscript{±0.314} & 61.80\textsubscript{±0.000} & 21.40\textsubscript{±0.000}         \\
    
& Self-Con  & \textbf{31.42}  &  32.87  & 34.50 & 31.88 & \textbf{68.78} & 53.50 &  62.69  & \textbf{24.50}        \\
& Step Back  & 31.43\textsubscript{±0.000} & 32.87\textsubscript{±0.322}  & --  &  31.85\textsubscript{±0.495} & 68.00\textsubscript{±0.000}   &  56.72\textsubscript{±0.000}        & --  & --        \\
    & SCoT 0-shot & \textbf{30.44}\textsubscript{±0.874} & \textbf{33.60}\textsubscript{±1.523} & \textbf{38.97}\textsubscript{±0.655} & \textbf{32.35}\textsubscript{±1.665} & \textbf{72.20}\textsubscript{±0.370} & \textbf{61.89}\textsubscript{±0.415} & \textbf{68.00}\textsubscript{±0.000} & \textbf{24.75}\textsubscript{±0.165}      \\ \cline{2-10}
                        
 & $\text{SCoT 1-shot}^-$  & 34.33\textsubscript{±0.000} & 31.50\textsubscript{±0.964} & 45.57\textsubscript{±1.087} & -- & 67.40\textsubscript{±0.000} & --  & -- & --        \\
& SCoT 1-shot  & \textbf{37.00}\textsubscript{±0.000} & \textbf{35.04}\textsubscript{±0.000}  & \textbf{47.38}\textsubscript{±0.107} & -- & \textbf{73.20}\textsubscript{±0.000} & -- & -- & --    \\ \hline  
\end{tabularx}
\caption{Accuracy (\%) using Llama2-8B and Mistral-7B across all datasets. $\text{SCoT 1-shot}^-$ refers to the results obtained using the standard few-shot CoT template but with demonstrations matched by strategy.}
\label{tab:Mainresult1full}
\end{table*}

%% file: appendixTables/all_results_3datasets_full.tex
\begin{table*}[ht]
\centering
\small
\begin{tabularx}{\textwidth}{cp{1.65cm}lllllllll}
\hline

Dataset &
  Method &
  Llama3-8B &
  Mistral-7b &
  Chatglm4-9B &
  Qwen2-7B &
  Qwen2-70B &
  Llama3.1-8B &
  Llama3.1-70B \\ \hline
  
\multirow{2}{*}{\begin{tabular}[c]{@{}c@{}}MMLU\\ Math\end{tabular}} &
  CoT 0-shot &
  46.67\textsubscript{±0.000} & 29.75\textsubscript{±0.924} & 66.67\textsubscript{±0.302} & \textbf{71.97}\textsubscript{±0.349}  & 84.20\textsubscript{±0.349}  & \textbf{59.63}\textsubscript{±0.000} & \textbf{85.19}\textsubscript{±0.605}   \\
 & SCoT 0-shot &
  \textbf{50.00}\textsubscript{±0.000}    & \textbf{32.35}\textsubscript{±1.665}   & \textbf{68.15}\textsubscript{±0.907} & \textbf{71.85}\textsubscript{±0.302} & \textbf{85.93}\textsubscript{±0.302} & 56.42\textsubscript{±0.175} & \textbf{85.19}\textsubscript{±0.000}  \\ \hline
                     
\multirow{2}{*}{SQA} & CoT 0-shot &
64.60\textsubscript{±0.595} & 56.22\textsubscript{±0.314} & 61.80\textsubscript{±0.363} & \textbf{61.00}\textsubscript{±0.000} & 75.22\textsubscript{±0.314} & 73.11\textsubscript{±0.314} & 64.67\textsubscript{±0.000} \\
                     & SCoT 0-shot      &
 \textbf{68.56}\textsubscript{±0.566} & \textbf{61.89}\textsubscript{±0.415} & \textbf{64.67}\textsubscript{±0.408} & \textbf{61.00}\textsubscript{±0.157} & \textbf{77.67}\textsubscript{±0.272} & \textbf{74.22}\textsubscript{±0.157} & \textbf{82.33}\textsubscript{±0.157}  \\ \hline
                     
\multirow{2}{*}{Object} &
  CoT 0-shot &
  44.27\textsubscript{±0.736} & 21.40\textsubscript{±0.000} & 61.80\textsubscript{±0.000} & 46.20\textsubscript{±0.000} & 93.93\textsubscript{±0.525} & 62.60\textsubscript{±0.411} & \textbf{100.00}\textsubscript{±0.000}   \\
                     & SCoT 0-shot       & 
  \textbf{68.40}\textsubscript{±0.000} & \textbf{24.67}\textsubscript{±0.000} & \textbf{69.00}\textsubscript{±0.000} & \textbf{47.53}\textsubscript{±0.094} & \textbf{97.47}\textsubscript{±0.339} & \textbf{77.60}\textsubscript{±0.993} & \textbf{100.00}\textsubscript{±0.000}  \\ \hline
\end{tabularx}
\caption{Accuracy(\%) across seven models on MMLU, SQA and Tracking\_Object datasets}
\label{tab:mainresult2full}
\end{table*}